\documentclass[lettersize,journal]{IEEEtran}
\usepackage{times}

\usepackage{multirow}
\usepackage{amsmath} 
\usepackage{multicol}
\usepackage{graphicx}
\usepackage{xspace}
\usepackage{xcolor}
\usepackage{caption}
\usepackage{wrapfig}
\usepackage{bbding}  
\usepackage{pifont}  
\usepackage{bbm} 
\usepackage{hyperref}
\usepackage[capitalise, nameinlink]{cleveref}

\usepackage{booktabs}
\usepackage{arydshln} 
\usepackage{siunitx} 
\usepackage{float}
\usepackage{amssymb}  
\usepackage{bm}
\usepackage[table]{xcolor}
\usepackage{dsfont}
\newcommand{\pingu}[0]{\textsc{PINGU}\xspace}

\newcommand{\levion}[0]{\textsc{Levion}\xspace}

\usepackage{siunitx}
\newcommand{\ci}[1]{%
    \textcolor{gray}{%
        \tiny~(\ensuremath{\pm \num{#1}})%
    }%
}
\usepackage{graphicx}

\usepackage[numbers]{natbib}

\newcommand{\ourrow}{\rowcolor{gray!7}}

\definecolor{citecolor}{HTML}{0099cc} 
\definecolor{lblue}{HTML}{ffb114} 
\definecolor{ogreen}{HTML}{2E7D32}
\definecolor{bred}{HTML}{BF360C}
\definecolor{newbrown}{HTML}{795548}

\hypersetup{
    colorlinks=true,
    linkcolor=citecolor,
    filecolor=magenta,      
    urlcolor=citecolor,
    citecolor=citecolor,
}

\usepackage{xfrac}

\usepackage{tikz}
\usetikzlibrary{positioning, fit, backgrounds, arrows.meta, calc}

\begin{document}

\title{\textbf{PINGU}: Extending Air-Bearing Spacecraft Emulators with Open-Source Actuators and Learned Control for Contact-Rich Proximity Operations}

\author{Ricard Marsal I Castan$^1$, Akiyoshi Uchida$^2$, Aman Arora$^1$, Pedro Lima$^1$, Matteo El-Hariry$^1$, Anrej Orsula$^1$, Francesco Grella$^1$, Antoine Richard$^1$, Cedric Pradalier$^3$, Miguel A. Olivarez-Mendez$^1$ \\$^1$University of Luxembourg, $^2$Tohoku University, $^3$Georgia Institute of Technology}

\twocolumn[{%
\renewcommand\twocolumn[1][]{#1}%
\maketitle
\vspace{-0.45cm}
\begin{center}
    \centering
    \captionsetup{type=figure}
     \includegraphics[width=1.0\textwidth]{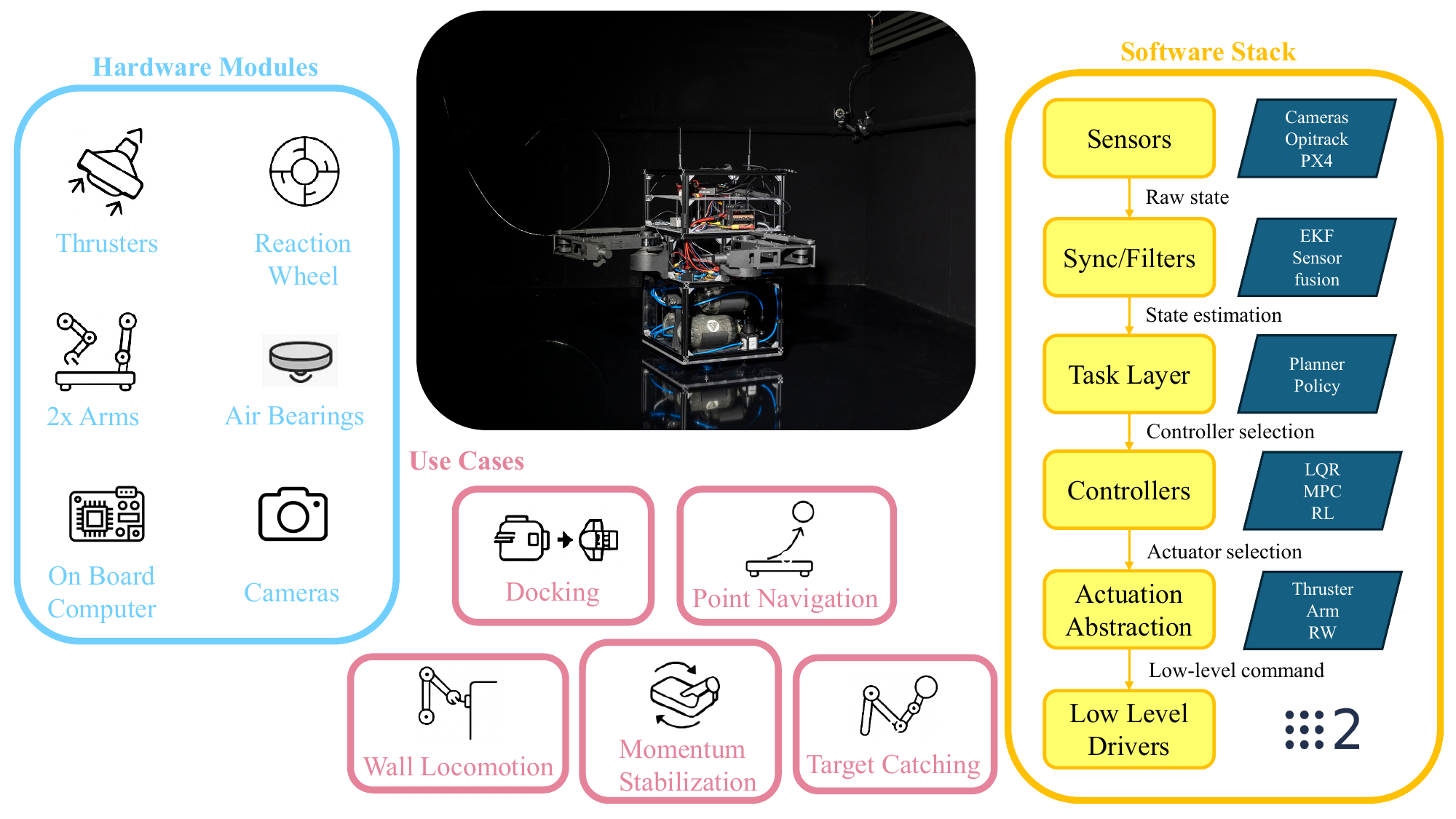}
     \vspace{-0.17in}
    \caption{\textbf{Overview.} Hardware modules (left) and the software stack (right), evaluated on \pingu, our ATMOS-compatible air-bearing platform integration, across the representative use cases shown (bottom).}
    \label{fig:pingu-overview}
\end{center}
\vspace{0.04in}
}]

\begin{abstract}
Low-cost planar air-bearing testbeds have matured into a standard proxy for free-flying spacecraft GNC, but they remain largely thruster-only and are rarely equipped for contact-rich, inertia-coupled manipulation. Building on the open-source ATMOS testbed, we contribute a reaction wheel and two force/torque-sensed robotic arms (LEVION) with interchangeable end-effectors, integrated as first-class control actuators through a unified ROS 2 abstraction layer. On top of the software stack we build a reinforcement-learning training environment and digital twin, and a controller that exploits these added degrees of freedom, letting classical optimal controllers and learned policies be swapped on the same hardware without modification. We validate the integrated system, PINGU, across four benchmark tasks: point-to-pose navigation (classical LQR vs. sim-to-real PPO), dynamic disturbance rejection under arm-induced center-of-mass shifts, reaction-wheel momentum stabilization, and force-controlled docking. The results show that these additions extend an ATMOS-class emulator into the contact-rich regime and bridge classical optimal control and reinforcement learning on one reproducible platform.
\end{abstract}
\IEEEpeerreviewmaketitle

\section{Introduction}
\label{sec:introduction}

On-orbit servicing, active debris removal, modular satellite assembly, and proximity operations all share a common technical core: vehicles that can simultaneously \emph{translate}, \emph{re-orient}, and \emph{physically interact} with other bodies in microgravity. As mission concepts move from rendezvous-and-dock toward dexterous on-orbit manipulation, the dynamics of the spacecraft itself become coupled to the motion of its on-board manipulators: every arm swing redistributes mass, perturbs the center of mass, and induces reaction torques that the attitude-control system must reject. Validating algorithms for this regime on flight hardware is prohibitively expensive, so the community has long relied on terrestrial \emph{floating platforms} that approximate frictionless planar motion using air bearings on a flat surface. In doing so they offer a low-cost, high-throughput proxy for the planar projection of orbital dynamics, and they have a long, well-surveyed history~\cite{Schwartz2003Historical,rybus2016planar}.

\paragraph{The limits of single-purpose testbeds}
Despite this maturity, the existing landscape of planar emulators is highly fragmented along functional lines. Thruster-only testbeds such as the Luleå and Luxembourg floating platform~\cite{banerjee2022floating, yalccin2023lightweight}, thruster-propeller testbeds like ATMOS~\cite{roque2025opensourcemodularspacesystems}, and thruster-and-reaction-wheel testbeds such as the ESA Orbital Robotics Lab platform~\cite{bredenbeck2022findingfollowingoptimaltrajectories}, M-STAR~\cite{Nakka2018ASD}, and ASTROS~\cite{Tsiotras2014ASTROS}, excel at studying guidance, navigation, and control (GNC) for free-flying base motion, but they do not carry on-board manipulators, so the coupled dynamics that arise from arm-induced inertia and CoM shifts cannot be reproduced. Conversely, the much shorter list of testbeds that \emph{do} integrate manipulators - for example York University's rendezvous-and-capture testbed~\cite{Santaguida2023}, the FIT Bob~\&~Charlie vehicles~\cite{Kwok-Choon2018Design}, and Sabatini~et~al.'s coordinated-manipulator platform~\cite{Sabatini2017Coordinated} - typically equip a single 3-DoF arm or passive grippers, lack co-integrated thrusters and reaction-wheel control authority, and are demonstrated under classical, hand-tuned controllers on narrow task suites. Astrobee~\cite{bualat2018astrobee,smith2016astrobee}, deployed on the ISS and used as a ground emulator, demonstrates the value of a unified flight-and-ground platform, but its single perching arm is not designed for the high-thrust, high-torque, contact-rich experiments that current robotic-servicing research demands. A recent analysis of intra-vehicular free-flyers~\cite{Turchetti2024IVFFS} echoes this conclusion at the system level: in-space manipulation with free-flying robots has been demonstrated only to a limited extent, and the dynamic challenges of a free-floating base coupled to an active manipulator remain a central, under-tested problem. A researcher today who wishes to ask, for example, \emph{``does a learned policy reject CoM shifts induced by arm motion better than an LQR with feed-forward inertia compensation?''} has no single platform on which to run both halves of the comparison cleanly.

\paragraph{Our approach}
We build directly on the open-source ATMOS testbed \cite{roque2025opensourcemodularspacesystems}, adopting much of its low-level software and a similar cold-gas thruster layout on a slightly restructured chassis, and extend it in three parts. First, we add two new control actuators — a high-torque reaction wheel for fine attitude control and momentum management, and two modular arms with 6-axis force/torque sensing and interchangeable end-effectors — exposed alongside the thrusters through a common ROS 2 actuator-abstraction layer, so that any controller (LQR, MPC, or a learned policy) can be swapped without changing the hardware interface. Second, on top of this stack we build a reinforcement-learning training environment and controller, and digital twin based on the RoboRAN \cite{elhariry2025roboran} and Space Robotics Bench simulator \cite{orsula2025spaceroboticsbench}, to exploit the additional degrees of freedom the wheel and arms provide. Third, we validate the integrated system in the lab as PINGU, \cref{fig:pingu-overview}, demonstrating the added actuators and the learned controller across four tasks: point-to-pose navigation, dynamic disturbance rejection under arm-induced CoM shifts, momentum stabilization, and force-controlled docking.

\paragraph{Bridging optimal control and reinforcement learning}
A second, complementary motivation underlies the platform's design. Spacecraft GNC has traditionally been the landscape of model-based optimal control, which provides strong theoretical guarantees but assumes a well-characterized rigid-body dynamics model, an assumption that is violated when on-board manipulators move and redistribute the body's mass. Reinforcement learning (RL) offers a path to controllers that adapt to such structured but hard-to-model disturbances, and recent work has begun to demonstrate sim-to-real RL on free-flying platforms~\cite{ElHariry2024DRIFT,Hovell2020DRL,Athauda2023RL} and on free-floating manipulators in simulation~\cite{Cao2023RLDualArm}. However, evaluation of RL on real hardware that simultaneously features thruster, reaction-wheel, \emph{and} manipulator authority (setting in which CoM coupling is most pronounced) remains absent. Sim-to-real transfer for free-flyers with reconfigurable inertia is therefore an open problem. Our actuators are designed to be a common substrate for both communities: every actuator is fully simulated in our training environment, every characterization curve is openly released, and the same task definitions and reward structures are exposed to classical and learned controllers alike. This enables direct comparisons of, for example, a tuned LQR baseline against a PPO policy with domain randomization, on the same physical robot performing the same task; a comparison that, to our knowledge, no existing planar testbed supports end-to-end on the full thruster+RW+arms suite.

\paragraph{Contributions}

\begin{itemize}
    \item \textbf{\pingu: platform with a set of openly released actuator modules} (a reaction wheel and two F/T-sensed robotic arms with interchangeable end-effectors) designed to co-integrate with the open-source ATMOS air-bearing testbed. We present the full mechanical and electronic design and a detailed characterization of every module (\cref{sec:system}), and position the resulting capabilities against the state of the art in \cref{tab:benchmark_comparison}.
    \item \textbf{A unified ROS\,2 control and estimation framework} (\cref{sec:system:software}) with an actuator-abstraction layer that exposes thrusters, reaction wheel, and arms through a common interface, allowing model-based (PID, LQR, MPC) and learned (PPO) controllers to be swapped without changes to the low-level stack.
    \item \textbf{Sim-to-real reinforcement-learning policies for dynamic disturbance rejection} under arm-induced center-of-mass and inertia shifts. We train PPO policies in IsaacLab \cite{mittal2025isaaclab} with domain randomization over base mass, center of mass, and a constant external wrench (a per-episode bias force and torque), and demonstrate transfer to hardware (\cref{sec:experiments}).
    \item \textbf{Experimental evaluation across four representative tasks}: point-to-pose navigation, dynamic disturbance rejection, momentum stabilization, and force-controlled docking; comparing classical and learned controllers on the same hardware (\cref{sec:experiments}).
    \item \textbf{Open-source release} of the full hardware design (CAD, BOM), firmware, ROS\,2 packages, simulation environments, training scripts, and experimental logs, to make \pingu reproducible and to lower the barrier of entry to learning-based spacecraft GNC research.
\end{itemize}

The rest of the paper is organized as follows. \cref{sec:related} positions \pingu against prior planar emulators, manipulator-equipped testbeds, and learning-based control on real robotic platforms. \cref{sec:system} describes the platform design and characterization, together with the unified ROS\,2 control and estimation framework. \cref{sec:experiments} reports the experimental evaluation across the four tasks. \cref{sec:discussion} and \cref{sec:conclusion} discuss lessons learned and future directions.

\section{Related Work}
\label{sec:related}

\begin{table*}[t]
    \centering
    \caption{Comparison of \pingu with existing planar/free-flying microgravity testbeds: ATMOS, Astrobee Flight (F), Astrobee Ground (G), Slider, ESA ORGL, Bob\,\&\,Charlie, M-STAR (6-DoF), JPL SSDT, the UniLu ZeroG floating platform, and our platform. Entries marked -- are not reported in the source paper.}
    \resizebox{\textwidth}{!}{%
    \begin{tabular}{lrrrrrrrrrr}
        \toprule
            & \textbf{ATMOS}~\cite{roque2025opensourcemodularspacesystems}
            & \textbf{Astrobee F}~\cite{bualat2018astrobee}
            & \textbf{Astrobee G}~\cite{smith2016astrobee}
            & \textbf{Slider}~\cite{banerjee2022floating}
            & \textbf{ESA}~\cite{bredenbeck2022findingfollowingoptimaltrajectories}
            & \textbf{Bob\,\&\,Charlie}~\cite{Kwok-Choon2018Design}
            & \textbf{M-STAR}~\cite{Nakka2018ASD}
            & \textbf{JPL SSDT}~\cite{wapman2021ssdt}
            & \textbf{ZeroG FP}~\cite{yalccin2023lightweight}
            & \textbf{\pingu (Ours)} \\
        \midrule
        \multicolumn{11}{l}{\textit{Physical properties}} \\
        \midrule
        \textbf{Mass [kg]}
            & 16.80 & 9.58 & 18.97 & 4.27 & 221.67 & 14.90 & --
            & -- & 5.32 & 25 \\
            
        \textbf{Moment of inertia [kg\,m$^2$]}
            & 0.297 & 0.162 & 0.252 & 0.190 & 12.223 & -- & $\mathrm{diag}(1.19,\,1.24,\,1.43)$
            & -- & 0.0591 & 0.72 - 1.35 \textsuperscript{c} \\
            
        \textbf{Height [m]}
            & -- & 0.32 & 0.32 & -- & 1.025 & -- & --
            & -- & 0.45 & 0.65 \\
            
        \textbf{Max. payload [kg]}
            & 150 & -- & -- & -- & -- & -- & --
            & -- & 20 & 1360 \\
        \midrule
        \multicolumn{11}{l}{\textit{Actuation}} \\
        \midrule
        \textbf{Num. thrusters}
            & 8 & 2 (impeller) & 8 & 8 & 8 & 8 & 16
            & 8 & 8 & 8 \\
        \textbf{Max. thrust per axis [N]}
            & 3.4 & 0.849 & 0.849 & 0.7 & 20.0 & -- & --
            & 0.51 & 1.0 & 1.5 \\
        \textbf{Max. torque [N\,m]}
            & 1.08 & 0.126 & 0.126 & -- & 1.5 & -- & --
            & -- & -- & 0.24 \\
        \textbf{Reaction wheel}
            & \ding{55} & \ding{55} & \ding{55} & \ding{55} & \ding{51} & \ding{51} & \ding{51}~(\,$\times$4)
            & \ding{55} & \ding{55} & \ding{51} \\
        \textbf{Integrated manipulators}
            & \ding{55}\,\textsuperscript{a} & 1 (perching) & 1 (perching) & \ding{55} & \ding{55} & gripper only & \ding{55}
            & \ding{55} & \ding{55} & \textbf{2 (F/T)} \\
            
        \textbf{F/T sensing on arm}
            & -- & \ding{55} & \ding{55} & -- & -- & \ding{55} & --
            & -- & -- & \ding{51} \\
        \midrule
        \multicolumn{11}{l}{\textit{Software \& reproducibility}} \\
        \midrule
        \textbf{Open-source HW/SW}
            & \ding{51} & partial & partial & \ding{55} & \ding{55} & \ding{55} & \ding{55}
            & \ding{55} & \ding{55} & \ding{51} \\
            
        \textbf{Learning-ready stack}
            & \ding{55} & \ding{55} & \ding{55} & \ding{55} & \ding{55} & \ding{55} & \ding{55}
            & \ding{55} & \ding{51}\,\textsuperscript{b} & \ding{51} \\
            
        \midrule
        \multicolumn{11}{l}{\textit{Demonstrated tasks}} \\
        \midrule
        \textbf{Pose tracking}
            & \ding{51} & \ding{51} & \ding{51} & \ding{51} & \ding{51} & \ding{51} & \ding{51}
            & \ding{51} & \ding{51} & \ding{51} \\
        \textbf{Docking / capture}
            & \ding{55} & \ding{51} & \ding{51} & \ding{55} & \ding{51} & \ding{51} & \ding{55}
            & \ding{55} & \ding{55} & \ding{51} \\
        \textbf{Disturbance rejection (shifting CoM)}
            & \ding{55} & \ding{55} & \ding{55} & \ding{55} & \ding{55} & \ding{55} & \ding{55}
            & \ding{55} & \ding{55} & \ding{51} \\
        \textbf{Contact-rich manipulation}
            & \ding{55} & \ding{55} & \ding{55} & \ding{55} & \ding{55} & \ding{55} & \ding{55}
            & \ding{55} & \ding{55} & \ding{51} \\
        \bottomrule
        \multicolumn{11}{l}{\footnotesize\textsuperscript{a}\,ATMOS demonstrates a manipulator \emph{payload} but does not include it as a characterized actuator in the base release.} \\
        \multicolumn{11}{l}{\footnotesize\textsuperscript{b}\,Via the external RANS/DRIFT toolchain~\cite{ElHariry2024DRIFT}, demonstrated for thruster-only control.}\\
        \multicolumn{11}{l}{\footnotesize\textsuperscript{c}\, Range depending on the arms configuration.}
    \end{tabular}%
    }
        \label{tab:benchmark_comparison}
\end{table*}

\pingu sits at the intersection of three threads in the literature: ground-based emulators of microgravity dynamics, on-board manipulation for free-floating spacecraft, and learning-based control for real robotic systems. We survey each thread in turn, then position \pingu against the closest competitors via the comparison in \cref{tab:benchmark_comparison}.

\subsection{Planar air-bearing emulators of microgravity}
\label{sec:related:emulators}

Air-bearing-supported floating platforms have served as mainstream terrestrial microgravity emulation for nearly half a century, and two surveys provide comprehensive historical context~\cite{Schwartz2003Historical,rybus2016planar}. The basic composition is invariant: a rigid body with porous-graphite or orifice-fed air bearings floats over an extremely flat surface made of granite, glass on optical bench, or epoxy resin, yielding three planar degrees of freedom (two translations, one rotation) with very low residual friction. Variations on this theme add hemispherical or spherical bearings to introduce limited rotational freedom, producing 5-DoF (e.g., ASTROS~\cite{Tsiotras2014ASTROS}) or 6-DoF (e.g., M-STAR~\cite{Nakka2018ASD}) facilities at the cost of mechanical complexity.

A representative cross-section of the modern planar testbed landscape illustrates the diversity of design choices. ATMOS~\cite{roque2025opensourcemodularspacesystems}, recently released as open-source by KTH, is built on a Pixhawk~6X and Jetson~Orin~NX stack, uses 8 PWM-driven cold-gas thrusters, and is intentionally compatible with the NASA Astrobee flight software. Bredenbeck et al.~\cite{bredenbeck2022findingfollowingoptimaltrajectories} describe a heavyweight ($221$ kg) modular stack at ESA's Orbital Robotics Lab, equipped with 8 solenoid thrusters and a reaction wheel, and demonstrate optimal trajectory tracking for the overactuated system; the related REACSA platform~\cite{bredenbeck2023reacsa} packages this thruster-plus-reaction-wheel actuation into a self-contained, controllable floating stack for orbital-robotics concept testing. The Luleå~\cite{banerjee2022floating} and Luxembourg~\cite{yalccin2023lightweight} floating platforms are, by contrast, lightweight ($4.27$ kg one and $5.0$ kg the other) thruster-only design controlled by nonlinear MPC, PID and RL-based policies. At the institutional end of the spectrum, NASA JPL's Small Satellite Dynamics Testbed~\cite{wapman2021ssdt} provides a rigorously characterized 8-thruster planar air-bearing platform whose propulsion-characterization methodology directly parallels our own (\cref{sec:system}). Other notable systems include the DLR TEAMS testbed for multi-spacecraft formation and contact-dynamics emulation~\cite{Schlotterer2010Testbed}, the Yonsei hardware-in-the-loop facility~\cite{Eun2018Development}, the NUAA SOOST testbed for on-orbit-operation simulation with vision-based navigation~\cite{Huang2022Characterizing}, the ESA ORBIT facility~\cite{Kolvenbach2014Orbit} (a $5\times9$~m epoxy floor with 14-camera VICON tracking), and the NTUA space-robot simulator~\cite{Papadopoulos2015NTUA}. These platforms have also become a proving ground for increasingly sophisticated control: online active fault estimation with collision avoidance on the Caltech testbed~\cite{Ragan2024OnlineTree}, and model-predictive schemes that handle the binary, dwell-time-constrained nature of solenoid thrusters together with a reaction wheel on the DFKI/ESA free-floating platform~\cite{stark2023linearmpc} (the same actuation challenge \pingu's base poses).

Across this body of work, two patterns can be observed. First, the prevailing actuation modality is thrusters, sometimes augmented by reaction wheels; on-board manipulators are the exception. Second, every system listed above evaluates a single controller class on a narrow task suite, making cross-platform comparison of \emph{control strategies} difficult. \pingu retains the design characteristics established by these platforms, like the open-source Pixhawk + Jetson stack of ATMOS, and broadens it along both axes by adding a reaction wheel and two F/T-sensed manipulators within a unified actuator-abstraction layer.

\subsection{Manipulation on free-floating ground emulators}
\label{sec:related:manipulation}

On-orbit servicing, active debris removal, and on-orbit assembly have driven decades of work on free-flying and free-floating space manipulators, surveyed comprehensively by Flores-Abad et al.~\cite{floresabad2014review} and, for the dynamics and control of the free-floating regime specifically, by Moosavian and Papadopoulos~\cite{moosavian2007freeflying}. The defining feature of this regime is that manipulator motion reacts back on an uncontrolled base, a coupling formalized in the foundational free-floating-manipulator literature via the generalized Jacobian's dependence on the (shifting) mass distribution~\cite{Umetani1989Resolved,Papadopoulos1991Dynamics}, and first demonstrated in orbit by the ETS-VII free-flying space robot~\cite{yoshida2003etsvii} and exploited operationally by the Orbital Express demonstration, which performed autonomous capture and robotic component transfer between two spacecraft~\cite{ogilvie2008orbitalexpress}. The DLR OOS-SIM facility~\cite{artigas2015oossim} emulates this servicer--target interaction on the ground using industrial robot arms in a hardware-in-the-loop scheme, complementing the air-bearing approach with full 6-DoF motion at the cost of true free-floating dynamics.

A smaller body of work integrates robotic manipulators directly onto air-bearing floating bases, motivated mainly by autonomous capture of free-floating targets. Sabatini et al.~\cite{Sabatini2017Coordinated} demonstrate coordinated base-and-arm maneuvers that exploit platform motion to optimize fuel efficiency, with simulation and experimental validation. Rybus et al.~\cite{rybus2019planar} reproduce the operations required for an orbital capture using a free-floating manipulator mounted on a planar air-bearing simulator. Santaguida and Zhu~\cite{Santaguida2023} present a planar testbed with a 3-DoF manipulator, a pseudo-galactic star tracker for pose estimation, and PD trajectory tracking, validated on a tumbling-target capture scenario. Nagaoka et al.~\cite{nagaoka2018dualarm} show that a \emph{dual}-arm space robot can capture and detumble a spinning target through repeated impacts, underscoring the value of two coordinated arms with contact sensing for capture --- a configuration \pingu realizes with full six-axis F/T feedback at each wrist. Bob~\&~Charlie~\cite{Kwok-Choon2018Design} are a pair of FIT air-bearing vehicles equipped with 8-thruster RCS, a reaction wheel, and various passive grippers and grasping features --- the closest existing system on actuator coverage, although their manipulation hardware is passive rather than active and the platform is demonstrated only at the open-loop characterization level. The Astrobee free-flyer~\cite{bualat2018astrobee,smith2016astrobee}, used both in flight aboard the ISS and as a ground emulator on a granite table, includes a single perching arm. A recent review of intra-vehicular free-flyers~\cite{Turchetti2024IVFFS} surveys SPHERES, Int-Ball, CIMON, and Astrobee, concluding that despite this lineage, in-space manipulation has been demonstrated only to a limited extent, and that the dynamic interaction between a free-floating base and an active manipulator remains a central challenge for autonomy.

The collective limitation of these systems is one of \emph{integration depth}: typically a single arm with limited sensing, no co-located reaction-wheel authority, and controllers tuned by hand for narrow tasks. None to our knowledge co-integrate dual F/T-sensed arms, a reaction wheel, and a thruster suite under a unified software stack designed to swap controllers without re-engineering the actuator interface.

\subsection{Reinforcement learning for spacecraft and floating-platform control}
\label{sec:related:rl}

Deep RL has been applied to a broad range of spacecraft GNC problems in simulation, from planetary landing~\cite{Gaudet2020DRLLanding} and asteroid proximity operations~\cite{gaudet2020terminal} to motion planning for dual-arm free-floating robots~\cite{Cao2023RLDualArm}, with a recurring emphasis on adapting to poorly modeled dynamics (e.g., domain-randomized attitude control across two orders of magnitude of mass--inertia variation~\cite{retagne2024adaptive}). Hardware demonstrations are narrower: sim-to-real PPO for pose and velocity tracking on a thruster-only floating platform (DRIFT~\cite{ElHariry2024DRIFT}), DRL guidance, proximity operations, and capture on the Carleton SPOT testbed~\cite{Hovell2020DRL,hovell2021deeprl,hovell2022capture}, collision-free docking~\cite{Athauda2023RL}, reaction-wheel attitude control under actuator failure~\cite{elhariry2025underactuated}, and, most recently, a free-flyer operating aboard the ISS~\cite{chapin2025apiary}. None of these combines thruster, reaction-wheel, and active-manipulator authority on one platform, leaving the most strongly inertia-coupled case untested on real hardware.

\pingu instead draws on the broader genre of \emph{research platforms purpose-built for learning-based control}, enabled by GPU-parallel simulation~\cite{makoviychuk2021isaacgym,rudin2022learning} and validated from legged robots~\cite{hwangbo2019learning,Liao2024BerkeleyHumanoid} to drone racing~\cite{kaufmann2023champion} and high-speed contact tasks~\cite{DAmbrosio2023TableTennis,Grandia2024Bipedal}. \pingu brings this methodology to planar spacecraft emulators through an open end-to-end release, a narrow sim-to-real gap, and classical and learned controllers evaluated on the same task interfaces. It complements space-focused simulation benchmarks~\cite{orsula2025spaceroboticsbench} with open hardware and focuses on reconfigurable inertia from on-board arms as the main learning challenge, which prior floating-platform RL work has not addressed.

\subsection{Positioning of \pingu}
\label{sec:related:positioning}

\cref{tab:benchmark_comparison} compares \pingu against representative prior platforms across actuation, sensing, software stack, and demonstrated tasks. Three observations follow. First, on \emph{actuator coverage}, \pingu is the only system surveyed to co-integrate a thruster suite, a reaction wheel, and two F/T-sensed arms with interchangeable end-effectors; the closest prior systems cover at most two of these categories (thrusters+RW for the ESA ORGL platform and M-STAR; thrusters plus a single arm for Santaguida~\&~Zhu~\cite{Santaguida2023} and Sabatini~et~al.~\cite{Sabatini2017Coordinated}; thrusters+RW with passive grippers for Bob~\&~Charlie). Second, on \emph{software stack}, \pingu extends the open Pixhawk\,+\,Jetson\,+\,ROS\,2 ecosystem shared with ATMOS by an actuator-abstraction layer spanning the heterogeneous arms+RW+thrusters mix, and is built from the outset as a substrate for both classical and learned controllers on identical task interfaces — whereas prior learning-on-hardware results in this class~\cite{ElHariry2024DRIFT} rely on external toolchains and thruster-only actuation. Third, on \emph{task coverage}, prior platforms typically demonstrate one or two task families; \pingu is evaluated on four (point-to-pose, disturbance rejection under shifting CoM, momentum stabilization, and force-controlled docking) with controllers swapped through a common interface, enabling a like-for-like classical-versus-learned comparison not previously reported for this platform class.

\section{System Design and Characterization}
\label{sec:system}

\begin{figure}[t]
    \centering
    \includegraphics[width=\linewidth]{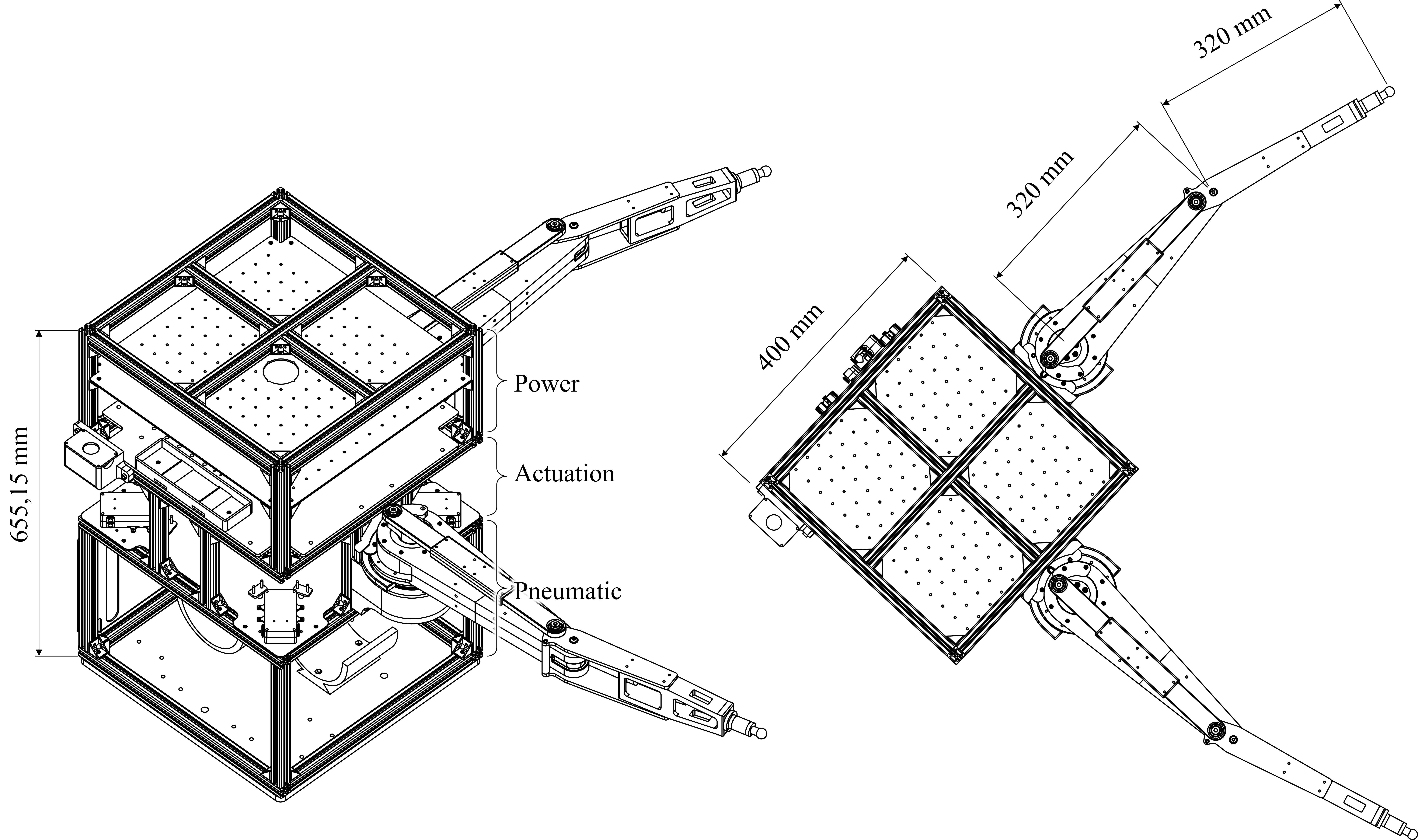}
    \caption{\textbf{\pingu system overview.}
    Annotated assembly drawing of the complete platform.
    The chassis is organized into four functional layers from bottom to top: pneumatic base (high-pressure gas bottles and regulator manifold), actuation mid-section (eight cold-gas thrusters, reaction wheel, and \levion mounts), power section (dual LiPo batteries and distribution electronics), and payload deck (exteroceptive sensor board and \levion).}
    \vspace{-1em}
    \label{fig:pingu_overview}
\end{figure}

\pingu is a reconfigurable planar microgravity emulator designed for the systematic evaluation of spacecraft GNC algorithms, with particular emphasis on contact-rich proximity operations and docking build on top of the open-source platform ATMOS\cite{roque2025opensourcemodularspacesystems}. The complete assembled platform is shown in \cref{fig:pingu_overview}. Three co-designed subsystems are described in this section: the mechanical chassis (\cref{sec:system:mechanical}), which houses the pneumatic supply, actuation modules, and payload interfaces; the electronics and sensing stack (\cref{sec:system:electronics}), which integrates flight-controller-grade avionics with a multi-modal exteroceptive sensor board; and the ROS\,2-based software design (\cref{sec:system:software}), which provides hardware abstraction and controller-deployment infrastructure.
Key actuation and sensing components are characterized experimentally alongside their hardware descriptions.

\subsection{Mechanical Architecture}
\label{sec:system:mechanical}

\pingu is built around a modular, vertically-layered chassis measuring $40 \text{cm} \times 40 \text{cm} \times 65 \text{cm}$ (\cref{fig:pingu_overview}). The structural skeleton consists of $20 \text{mm} \times 20 \text{mm}$ aluminum T-slot extrusion profiles joined by polycarbonate side panels, while non-load-bearing enclosures, brackets, and arm linkages are fabricated via 3D printing. The chassis is organized into four stacked functional sections along its vertical axis: a pneumatic base containing the pressure bottles and regulator manifold, an actuation mid-section housing the thruster valves, reaction wheel, and robotic arm mounts, a power section with the battery packs and power-distribution electronics, and an open payload deck rated. The pneumatic supply, actuator specifications, and robotic arm configuration are detailed in the following subsections.

\subsubsection{Pneumatic Section}
Like the open-source ATMOS platform~\cite{roque2025opensourcemodularspacesystems}, \pingu uses three high-pressure gas bottles (1.5 L each, filled to 200 bar) routed through a regulator manifold on the pneumatic base plate.
The manifold steps the supply down to 5 bar on two separate regulated rails: one for the three air bearings and one for the solenoid thrusters.

\begin{figure}[t]
    \centering
    \includegraphics[width=0.55\linewidth]{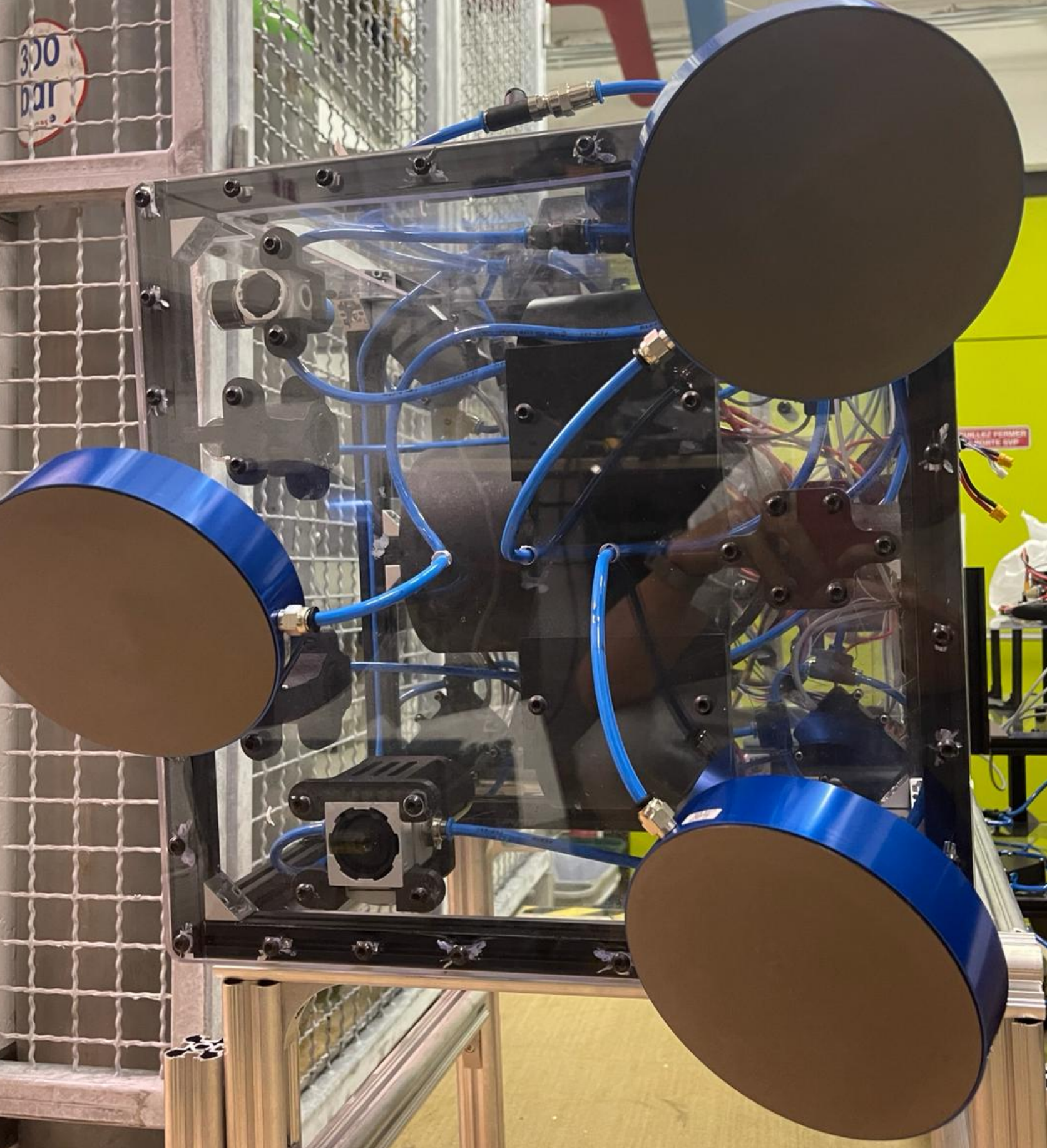}
    \caption{\textbf{New~Way 150 mm flat round air bearing (S1015001).}
    Each bearing sustains a 6~\si{\micro\metre} aerostatic gap at 5 bar and is rated for 453.5 kg static load.
    Three bearings are arranged in an equilateral triangle beneath \pingu's base plate, providing near-zero friction over the test floor.}
    \vspace{-1em}
    \label{fig:airbearing}
\end{figure}

Frictionless planar motion is provided by three New~Way \cite{newway150mmairbearing} 150 mm flat round porous-graphite air bearings (\cref{fig:airbearing}), arranged in an equilateral triangle beneath the base plate.
Compared to the 60 mm bearings used on ATMOS, the 150 mm format substantially increases load capacity per bearing: at the nominal 5 bar supply (within the rated 4.1 - 5.5 bar range) each bearing sustains an aerostatic gap of approximately 6~\si{\micro\metre} at a flow of 2.9 - 3.5 NLPM and is rated for an ideal static load of 453.5 kg.
The three-bearing arrangement yields a combined bearing-limited load capacity of approximately 1360 kg, supporting a significantly heavier platform with heavier payloads, such as the \levion and their associated hardware.

\subsubsection{Actuation Modules}
\pingu employs three complementary actuation modalities: eight cold-gas thrusters for impulsive planar force generation, a reaction wheel for continuous torque control, and two \levion arms for inertia reconfiguration and contact manipulation.
Together they cover the full range of free-floating spacecraft maneuvers within the planar microgravity environment.

\begin{figure}[h]
    \centering
    \includegraphics[width=\linewidth]{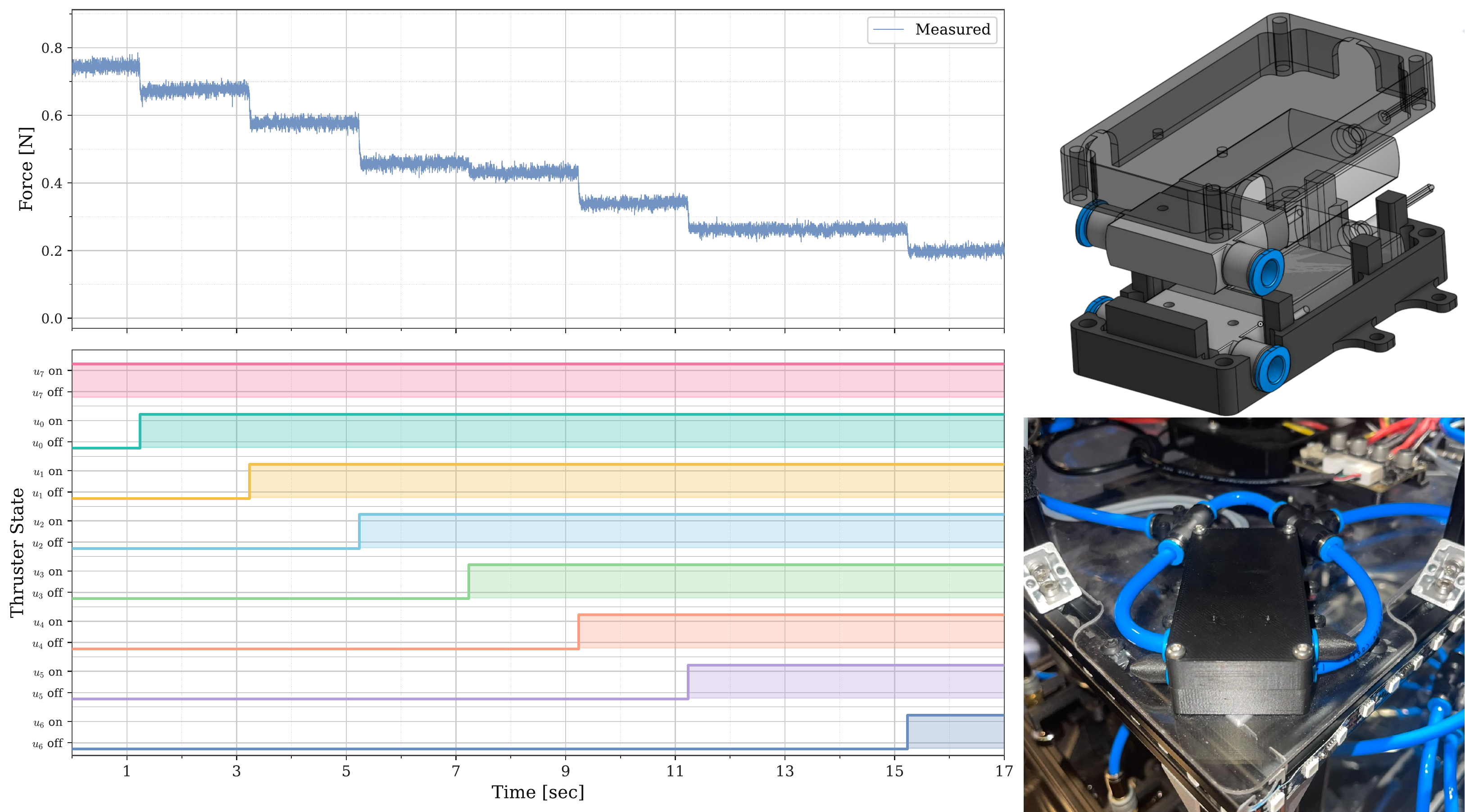}
    \caption{\textbf{Thruster characterization.}
    \emph{Left:} Measured force output as thrusters are opened sequentially (1 through 8), showing the progressive pressure drop across the shared supply manifold that reduces individual thrust output under concurrent actuation.
    \emph{Right:} CAD model of the thruster box alongside the assembled unit mounted on the \pingu actuation mid-section.}
    \vspace{-1em}
    \label{fig:thruster-charact}
\end{figure}

\paragraph{Thrusters}
The eight thrusters function analogously to a spacecraft's Reaction Control System (RCS): solenoid valves with a maximum switching frequency of 500 Hz, operated in software at a fixed carrier frequency of 10 Hz with a pulse width ranging from 0 ms to 100 ms (full off to full on) for proportional thrust modulation.
Air is supplied at a manifold pressure of 5 bar, drawn through 4 mm tubing and expelled via 2 mm nozzles.
The thrusters are arranged in a quad-corner configuration: two at each corner of the actuation mid-section with nozzle axes in mutually orthogonal body-frame directions ($\hat{x}$ and $\hat{y}$), providing independent force generation along both planar axes and torque about the yaw axis.
Isolated characterization, performed by firing a single valve into a force plate, yields approximately 0.7 N per thruster.
Because all eight valves share a common supply manifold, simultaneous actuation causes a pressure drop that reduces individual thrust output; the resulting coupling is characterized in \cref{fig:thruster-charact} and must be accounted for in any thrust-allocation model.

\begin{figure}[b]
    \centering
    \includegraphics[width=\linewidth]{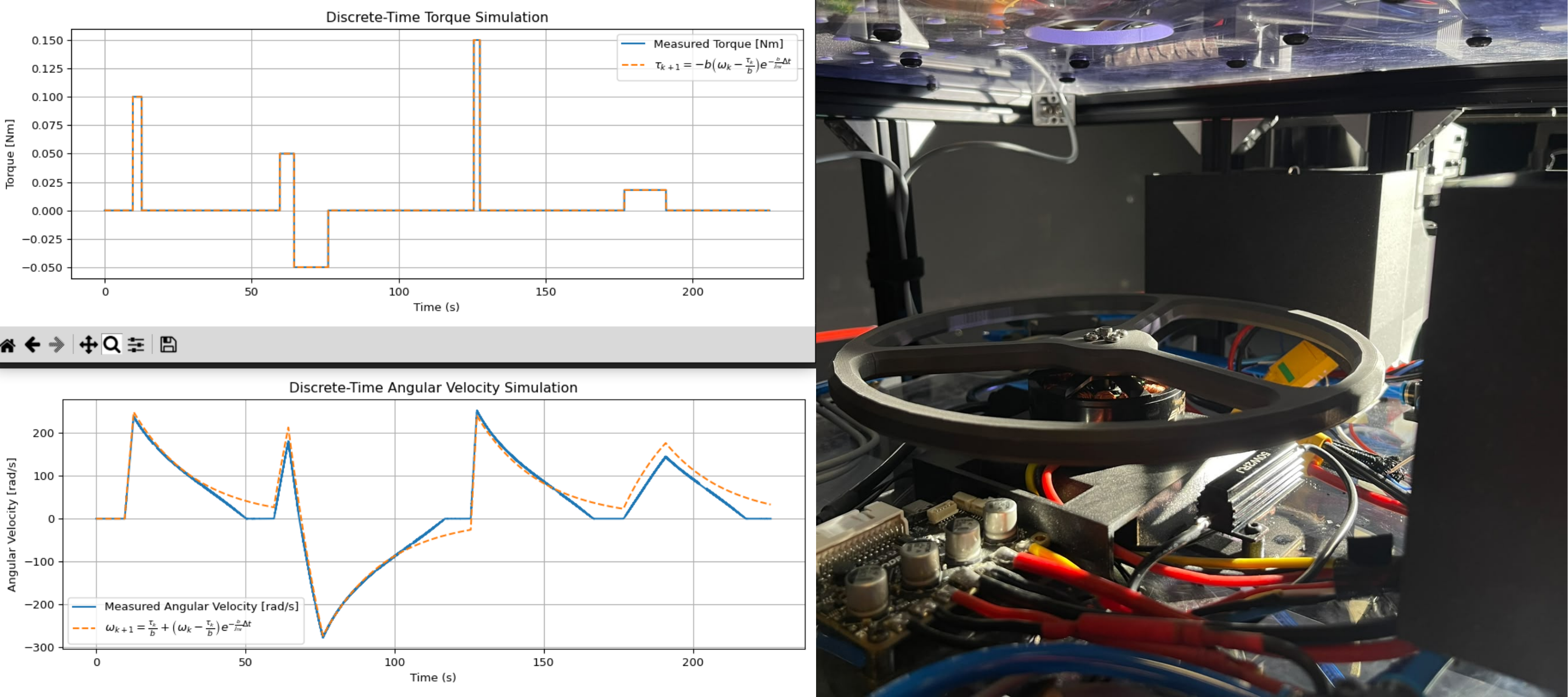}
    \caption{\textbf{Reaction wheel.}
    (A)~Measured torque-speed envelope of the reaction wheel assembly, showing usable attitude-control authority across the operating speed range. (B)~The 20 cm metal-blended PLA disk mounted on the D5312S dual-shaft motor inside the \pingu actuation mid-section.
    }
    \label{fig:rw-charact}
\end{figure}

\paragraph{Reaction wheel}
A single reaction wheel provides continuous, non-impulsive attitude authority, complementing the binary solenoid thrusters which cannot produce smooth yaw torque profiles.
The wheel is a 20 cm diameter disk 3D-printed in metal-blended PLA to maximise rotational inertia at low mass; the open-source design can equally be CNC-machined from any solid material for higher inertia requirements.
It is driven by a D5312S 330\,KV dual-shaft brushless motor controlled by an ODrive S1 \cite{odriverobotics}, with the disk mounted on one output shaft and an AMT212B magnetic absolute encoder on the other.
The motor delivers a torque constant of 0.025 $\frac{Nm}{A}$.
The reaction wheel angular velocity $\Omega_{r}$ and generated torque $\tau_{r}$ are modelled in discrete time as  $$\Omega_{r_{k+1}} = \frac{\tau_{cmd_{k}}}{b} + \left(\Omega_{r_{k}} - \frac{\tau_{cmd_{k}}}{b}\right)\exp{\left(-\frac{b}{J_r}\Delta t\right)}, \text{and}$$ $$\tau_{r_{k+1}} = J_r\dot{\Omega}_{r_{k+1}} = \left(\tau_{cmd_{k}} - b\Omega_{r_{k}}\right)\exp{\left(-\frac{b}{J_r}\Delta t\right)}$$ respectively, where $b$ is a viscous friction constant, $J_r$ is the wheel moment of inertia about its spin axis, and $\tau_{cmd}$ is the command torque at a given step.
\cref{fig:rw-charact} show a comparison between the modelled wheel dynamics and the measured torque-speed envelope, across the full operating speed range.

\paragraph{LevionArms}
The \levion are the most distinctive element of \pingu: a pair of planar 2-DOF robotic arms that serve a dual role not found in existing air-bearing testbeds.
First, by repositioning their links, they actively redistribute the platform's effective centre of mass and moment of inertia, enabling systematic evaluation of controller robustness across different inertial regimes without changing the hardware.
Second, they act as contact-manipulation end-effectors for proximity operations and docking, exercising the full sensing and actuation stack in a single closed-loop task.

\begin{figure}[t]
    \centering
    \includegraphics[width=\linewidth]{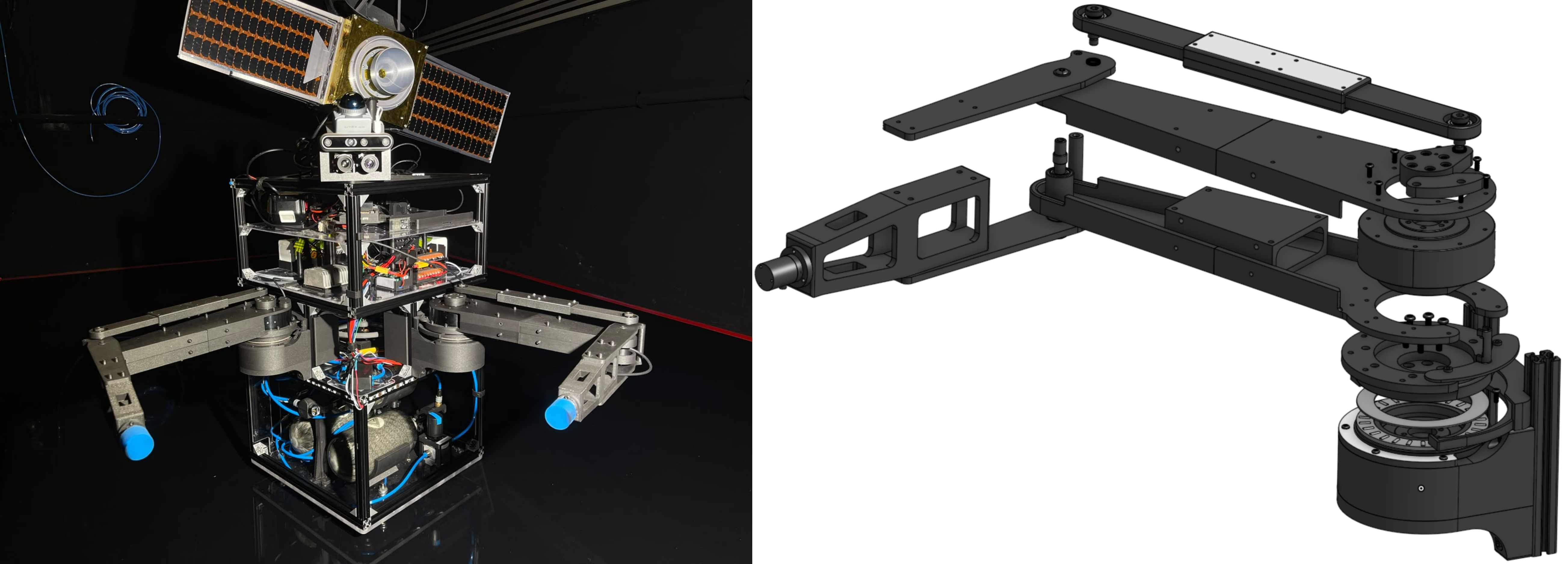}
    \caption{\textbf{\levion design.}
    Planar 2-DOF arm mounted on \pingu's actuation mid-section.
    Each joint is driven by a CubeMars AK80-8 KV60 quasi-direct-drive motor; a Leptrino six-axis F/T sensor at the wrist provides wrench feedback for impedance control.
    The symmetric bilateral mounting allows the arms to reconfigure the platform's effective inertia distribution on-the-fly.}
    \vspace{-1em}
    \label{fig:levionArm}
\end{figure}

Each arm forms a shoulder-elbow serial chain (\cref{fig:levionArm}) with a total mass of 2.28 kg, with links 3D-printed for low mass and rapid reconfiguration.
Both joints are actuated by CubeMars AK80-8 KV60 quasi-direct-drive actuators (rated output torque: 10 Nm; peak output torque: 25 Nm; 8:1 reduction ratio). During the experiments, the commanded joint velocity and torque were limited to 10 rad/s and 10 Nm, respectively; the low gear ratio maximizes backdrivability and torque transparency, which is critical for compliant contact.
The shoulder joints share a symmetric range of $\pm 1.56$ rad; the elbow joints are geometrically mirrored --- left elbow $[-2.61,\,-0.36]$\,rad, right elbow $[0.36,\,2.61]$\,rad --- yielding a manipulation workspace symmetric about the platform centreline. Each arm's motors are torque-controlled through a PD law that allows compliant position reference tracking.
A Leptrino six-axis F/T sensor at each wrist provides full wrench feedback for impedance control and contact-onset detection.
A standardized wrist coupling enables rapid interchange between grippers, suction cups, and custom tools without any structural modification.

\subsubsection{Payload}
\pingu rests on three New~Way 150 mm flat round air bearings, each rated at 453.5 kg, giving a combined bearing-limited load capacity of 1360 kg; in practice, the usable payload budget is constrained by the platform's own mass and the laboratory air supply.
The mid-deck section provides side mounts for the two \levion arms, leaving the top deck as a dedicated, highly customizable payload area. In the default configuration, this top deck carries the exteroceptive sensor board. The entire mounting interface is designed to easily accommodate alternative actuation modules, sensor arrays, or specialized experimental hardware, allowing the platform to be reconfigured for different tasks and weight-distribution requirements without structural modifications.

\subsection{Electronics and Sensors}
\label{sec:system:electronics}

\pingu's electronics subsystem integrates a flight-grade compute stack, a shared CAN actuator bus, and a dual-battery power distribution network, together with a dedicated exteroceptive sensor payload described in \cref{sec:system:sensors-board}.

\subsubsection{Electronics}
The central compute stack is the Holybro Pixhawk Jetson Baseboard, which integrates a Pixhawk 6X flight controller and an NVIDIA Jetson Orin NX (16~GB RAM) into a single package, simplifying the hardware and software integration of PX4 with a companion computer. The Pixhawk and Jetson communicate via the board's internal Gigabit Ethernet switch, which carries both the ROS\,2 inter-process traffic and the micro-XRCE-DDS bridge between PX4 and the companion computer.

\paragraph{Actuator drivers and CAN bus}
The reaction wheel motor and all four joints of the two \levion arms share a single CAN bus routed through the actuation mid-section. The reaction wheel is driven by an ODrive S1 motor driver; each arm joint is driven by the integrated CAN-FD interface of the CubeMars AK80-8 KV60 actuator. All five joints are exposed to ROS\,2 via a \texttt{ros2\_control} hardware interface running at 200 Hz, providing a hardware-agnostic actuation API to the software stack.

\begin{figure}[t]
    \centering
    \includegraphics[width=\linewidth]{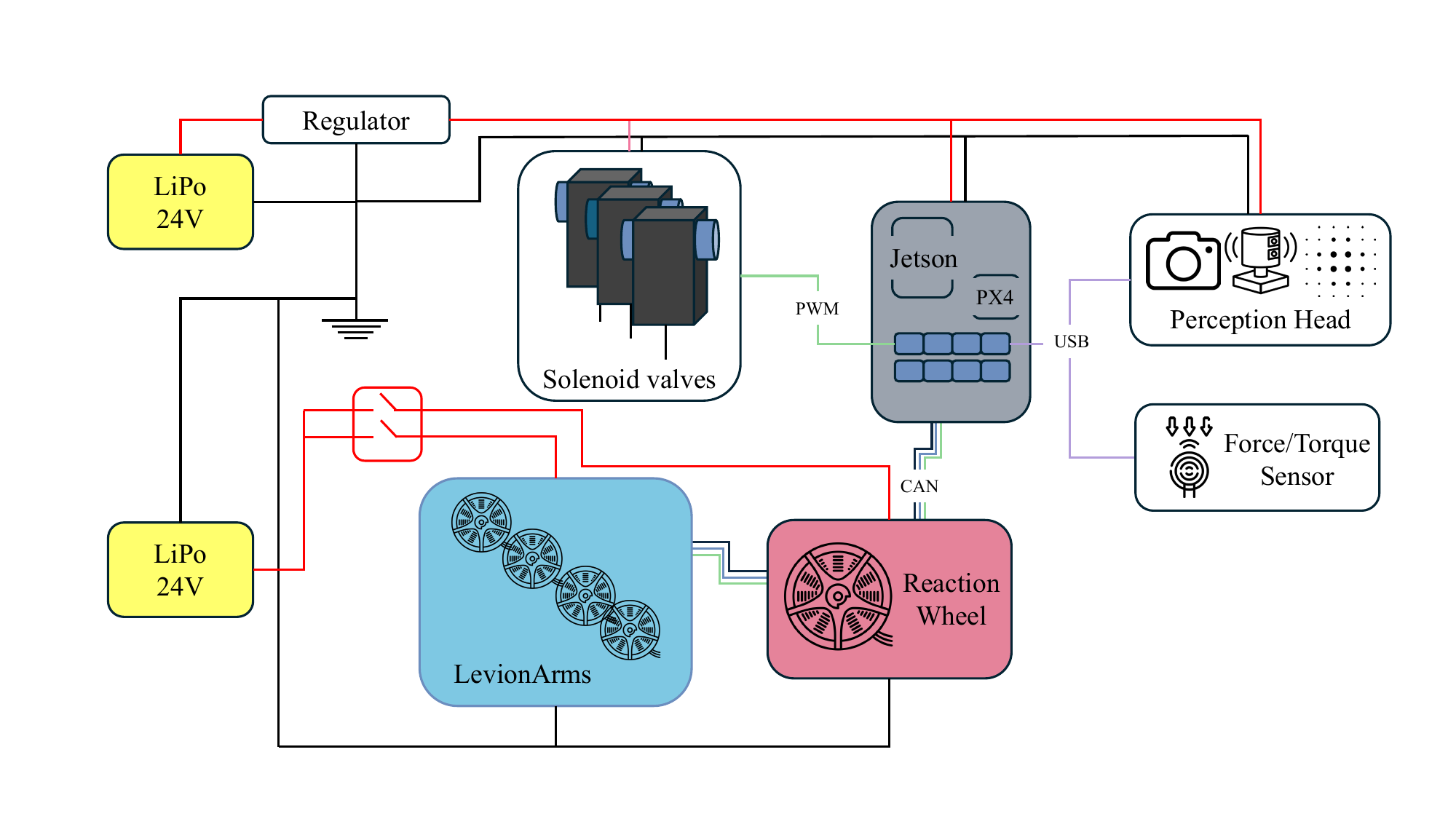}
    \caption{\textbf{Electronics overview.} Two-domain power architecture of \pingu. The control domain (Battery~1, 22.2~V nominal, regulated) supplies the flight controller, compute, and solenoid loads. The actuation domain (Battery~2, direct) supplies the ODrive S1 and CAN-FD arm actuators. Signal paths -- including CAN bus, micro-XRCE-DDS bridge, and Ethernet -- are shown alongside the power rails. Both domains share a common ground.}
    \vspace{-1em}
    \label{fig:electronics-overview}
\end{figure}

\paragraph{Power distribution}
The system is powered by two 6S 9.5 Ah LiPo batteries (22.2 V nominal) with a shared ground, partitioned into two isolated domains to prevent high-current actuation transients from disturbing the control electronics (\cref{fig:electronics-overview}).
The \emph{control domain} draws from the first battery through a regulated 24 V buck-boost converter, supplying the Pixhawk~6C, Jetson Orin NX, sensor board, and thruster solenoids.
The \emph{actuation domain} draws from the second battery directly, powering the ODrive S1 motor controller and all four CAN-FD arm actuators.

\subsubsection{Sensor Suite}
\label{sec:system:sensors-board}

\begin{figure}[t]
    \centering
    \includegraphics[width=\linewidth]{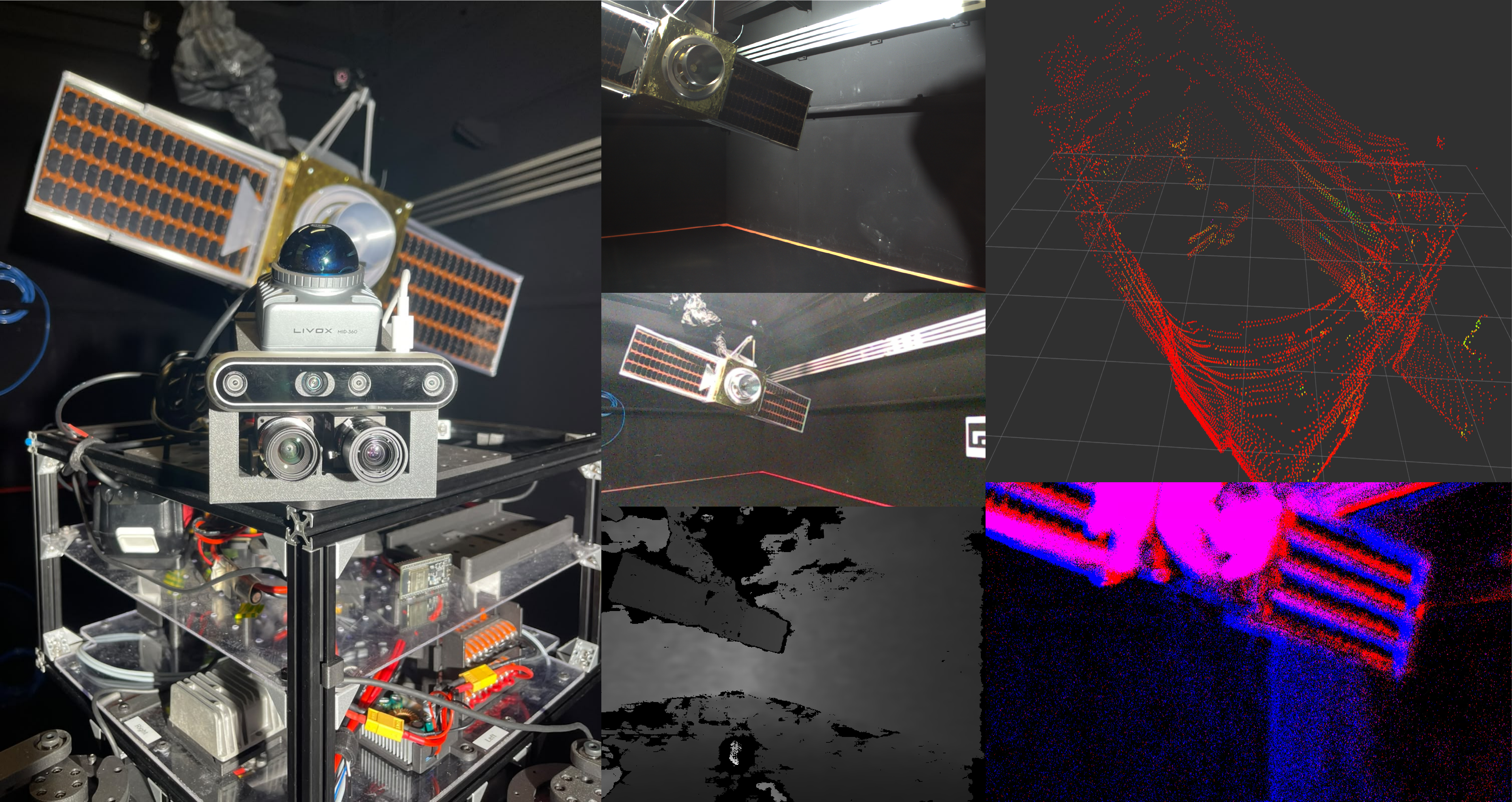}
    \caption{\textbf{\pingu sensor suite.} The four exteroceptive modalities mounted on the dedicated sensor board (Intel RealSense D455, FLIR Firefly, Prophesee event camera, Livox LiDAR), together with representative output from each.}
    \vspace{-1em}
    \label{fig:sensor_stack}
\end{figure}

\pingu's sensing is organized across three complementary layers.
Proprioceptive state is provided by the Pixhawk~6C flight controller, which exposes a 3-axis IMU (accelerometer and gyroscope), barometer, and magnetometer as standard ROS\,2 sensor topics via the micro-XRCE-DDS bridge, making inertial measurements available to all control nodes without additional drivers.
Each \levion arm wrist embeds a 6-axis force/torque sensor whose readings are surfaced through the \texttt{ros2\_control} hardware interface, providing direct contact sensing for manipulation and docking tasks.
Exteroceptive perception is offloaded to a dedicated sensor board based on the NVIDIA Jetson AGX Orin Developer Kit, mounted on \pingu's payload deck; all sensor drivers run on this board and publish data as ROS\,2 topics available to the main Jetson Orin NX over the onboard Ethernet network (\cref{fig:sensor_stack}).

The exteroceptive suite comprises four complementary modalities.
An Intel RealSense D455 RGB-D camera provides colour imagery and depth point clouds via an active infrared stereo pair, primarily used for close-range target detection and docking.
A FLIR Firefly RGB monocular camera provides a high-frame-rate colour stream.
A Prophesee event camera offers sub-millisecond-latency asynchronous brightness-change detection, enabling high-speed proximity sensing robust to motion blur.
A Livox mid-range LiDAR provides dense 3-D point clouds for environment mapping and obstacle avoidance.
Together, the four modalities span the full spectrum from high-frequency, low-latency event data to dense geometry, allowing sensing pipelines to be benchmarked on the same platform and task suite.

\subsection{Software Design}
\label{sec:system:software}

\begin{figure}[t]
    \centering
    \includegraphics[width=\linewidth]{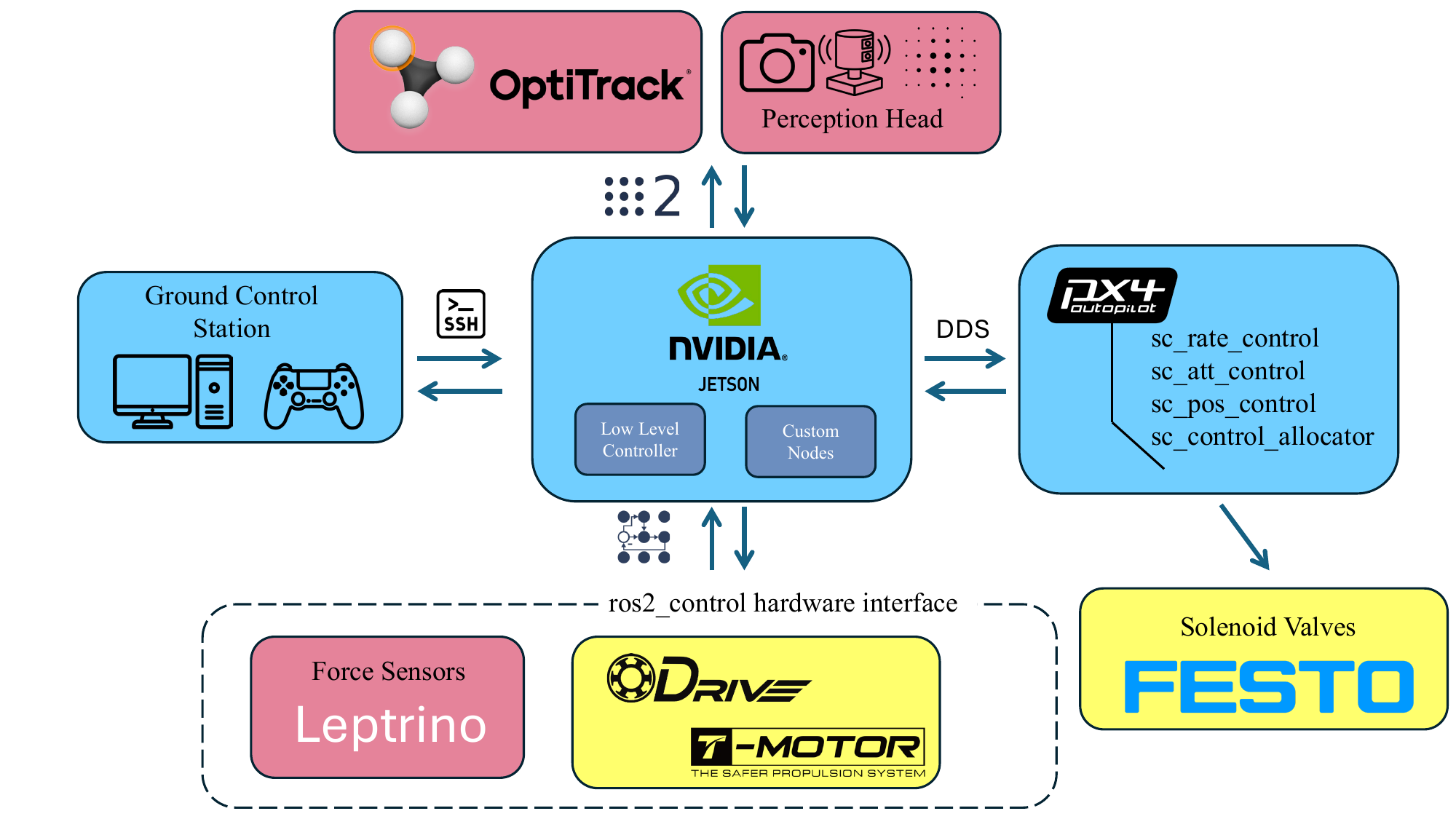}
    \caption{Two-workspace ROS\,2 software architecture of \pingu: the low-level platform workspace exposes all actuators as standard interfaces; the controller-deployment workspace implements classical and learned control logic above that abstraction boundary.}
    \vspace{-1em}
    \label{fig:software-architecture}
\end{figure}

\pingu's software is organized into two cooperating ROS\,2 workspaces that cleanly separate hardware abstraction from control logic (\cref{fig:software-architecture}). The low-level platform workspace runs on the Jetson Orin NX and is responsible for exposing every actuator as a standard ROS\,2 interface. The controller-deployment workspace runs on the same machine and implements all control and inference logic. This separation allows controllers (classical or learned) to be developed, evaluated, and swapped without any modification to the hardware-facing layer.

\paragraph{Low-level hardware abstraction}
All actuators are exposed through two complementary mechanisms. Thruster commands are handled by a dedicated PX4 offboard node that places the Pixhawk 6X into direct-actuator offboard mode via the micro-XRCE-DDS bridge and forwards an 8-element command vector as an \texttt{ActuatorMotors} uORB message over the DDS bus. The robotic arms and reaction wheel are managed by a \texttt{ros2\_control} hardware interface that communicates with the CAN bus at $200$ Hz, exposing position, velocity, and effort command interfaces for all five actuated joints. A command multiplexer node arbitrates between concurrent command sources, allowing a user to manually control the platform through a joystick or let the autonomous controllers seamlessly take over, while enforcing priority ordering and safety limits across the two actuation paths.
Sensor data enters the ROS\,2 graph through two drivers: a VRPN client publishes OptiTrack motion-capture poses at 100 Hz, and a RealSense driver streams RGB-D frames for vision-based tasks.

\begin{figure}[t]
    \centering
    \includegraphics[width=\linewidth]{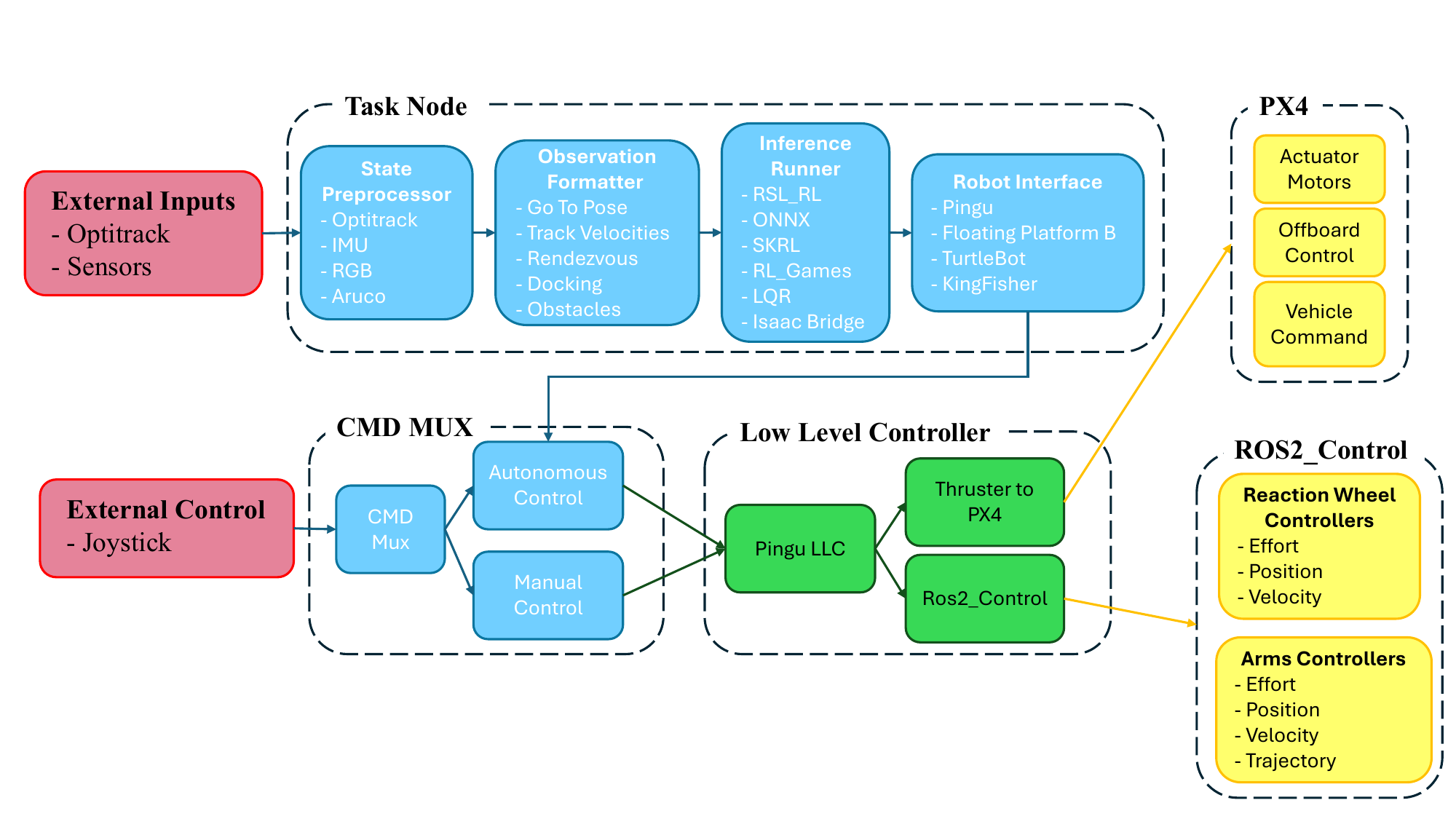}
    \caption{\textbf{ROS\,2 deployment node graph.}
    External inputs (red) feed the \emph{Task Node}, which chains four
    interchangeable modules (blue): a \emph{state preprocessor} (sensor
    drivers and velocity estimation), \emph{observation formatter}
    (task-specific input assembly), \emph{inference runner} (policy
    evaluation), and \emph{robot interface} (action mapping).
    The robot interface publishes a unified command to the \emph{CMD~Mux},
    which arbitrates between autonomous and manual (joystick) control sources.
    The \emph{Low-Level Controller} (green) routes thruster commands to PX4
    via an \texttt{ActuatorMotors} uORB message and joint commands to \texttt{ros2\_control}
    (yellow), which manages effort, position, velocity, and trajectory
    controllers for the reaction wheel and both \levion arms over the CAN bus.
    Every module is hot-swappable at launch time; the hardware layer
    requires no modification when the task, sensor, or policy changes.}
    \vspace{-2em}
    \label{fig:ros2-nodes}
\end{figure}

\paragraph{Controller deployment and inference}
All controllers share a common output interface: a 15-element command vector
comprising a bearing flag, a thruster-enable bit, eight binary thruster
duty-cycle values, four \levion arm joint setpoints, and one reaction-wheel
torque command.
Classical controllers are implemented as ROS\,2 nodes that subscribe to the
state topics and publish to this interface directly.
Learned policies are deployed via a ROS\,2 node
built on the RoboRAN framework~\cite{elhariry2025roboran}, which runs a fixed-rate
inference loop at 10 Hz (\cref{fig:ros2-nodes}).
The node composes four interchangeable modules in a fixed pipeline.
A \emph{state preprocessor} converts raw sensor data into the body-frame
state required by the policy; the default implementation buffers
OptiTrack pose messages at 100 Hz over a 30-frame window and
numerically differentiates them to obtain linear and angular velocities,
while drop-in alternatives (ArUco marker detection, odometry) are
registered via the same factory interface.
A task-specific \emph{observation formatter} assembles the policy input
tensor from the preprocessed state and a received goal; for example,
the \emph{GoToPose formatter} computes goal distance, heading errors
(sine/cosine encoded), body-frame velocities, and the previous action vector.
An \emph{inference runner} evaluates the loaded policy at each step. The platform natively supports RSL-RL \cite{schwarke2025rslrl} checkpoints, executing both MLP and GRU architectures while directly restoring empirical observation-normalization statistics. For broader compatibility, the runner also integrates ONNX Runtime (leveraging GPU acceleration via NVIDIA TensorRT on the Jetson), SKRL, RL-Games, and a classical LQR baseline. A \emph{robot interface} maps the policy output to the
shared 15-element command vector.
A simulation bridge is also provided, allowing a policy to step inside
IsaacLab \cite{mittal2025isaaclab} while issuing live actions to the hardware, enabling closed-loop
validation before committing to fully off-policy deployment.
Switching between a classical and a learned controller requires only
relaunching the controller-layer nodes; the hardware-abstraction layer
and all sensor pipelines remain unchanged.

\begin{figure}[t]
    \centering
    \includegraphics[width=\linewidth]{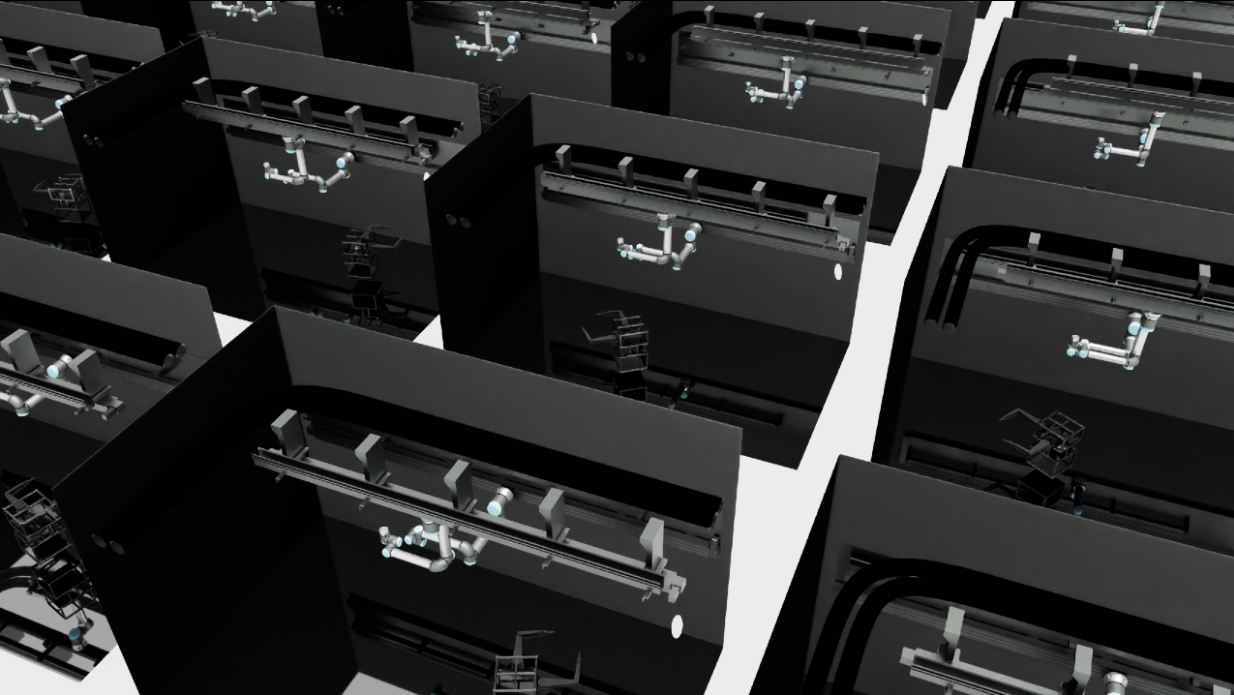}
    \caption{\textbf{\pingu digital twin in the Space Robotics Bench.} Thousands of parallelized copies of the laboratory environment running simultaneously on a single GPU inside IsaacLab. Each instance contains a full \pingu model (air-bearing platform with two \levion arms) and two UR10 manipulator mounted to the ceiling and wall rails, replicating the physical laboratory layout. Massively parallel rollouts with domain randomization enable the sim-to-real transfer results reported in \cref{sec:experiments}.}
    \vspace{-1em}
    \label{fig:digital-twin}
\end{figure}

\paragraph{Digital Twin}
The \pingu digital twin is implemented within the Space Robotics Bench (SRB)~\cite{orsula2025spaceroboticsbench}, a GPU-accelerated robot-learning framework built on top of IsaacLab~\cite{mittal2023orbit} and NVIDIA Isaac Sim. The simulated environment faithfully reproduces the full platform dynamics — rigid-body kinematics, binary solenoid-valve thrusters, reaction wheel, and 2-DOF \levion arms with configurable end-effectors — as well as the physical laboratory layout, including the precision-levelled air-bearing test floor, an OptiTrack motion-capture system, and two ceiling-mounted Universal Robots UR10 manipulators that serve as interactive targets for proximity-operations and contact-dynamics tasks (\cref{fig:digital-twin}). Exploiting PhysX rigid-body simulation, SRB runs thousands of environment instances in parallel on a single GPU, enabling sample-efficient RL training with domain randomization. Beyond training, the digital twin is used for controller validation before hardware deployment and as a real-time state-visualization interface during laboratory experiments.

\section{Experimental Evaluation}
\label{sec:experiments}

The experiments presented here are designed to demonstrate the breadth of \pingu's capabilities rather than to benchmark algorithms in isolation. Concretely, we aim to show three things: \emph{(i)} the platform's unified actuator-abstraction layer allows both classical optimal controllers and end-to-end learned policies to be deployed on the same hardware, on the same tasks, without modifying the underlying stack; \emph{(ii)} the platform's modular sensor suite can be integrated directly into the control pipeline, enabling sensor-driven tasks beyond blind pose tracking; and \emph{(iii)} the digital twin is a faithful enough model of the real platform that policies trained entirely in simulation transfer to hardware with acceptable performance.

\begin{figure}[t]
    \centering
    \includegraphics[width=\linewidth]{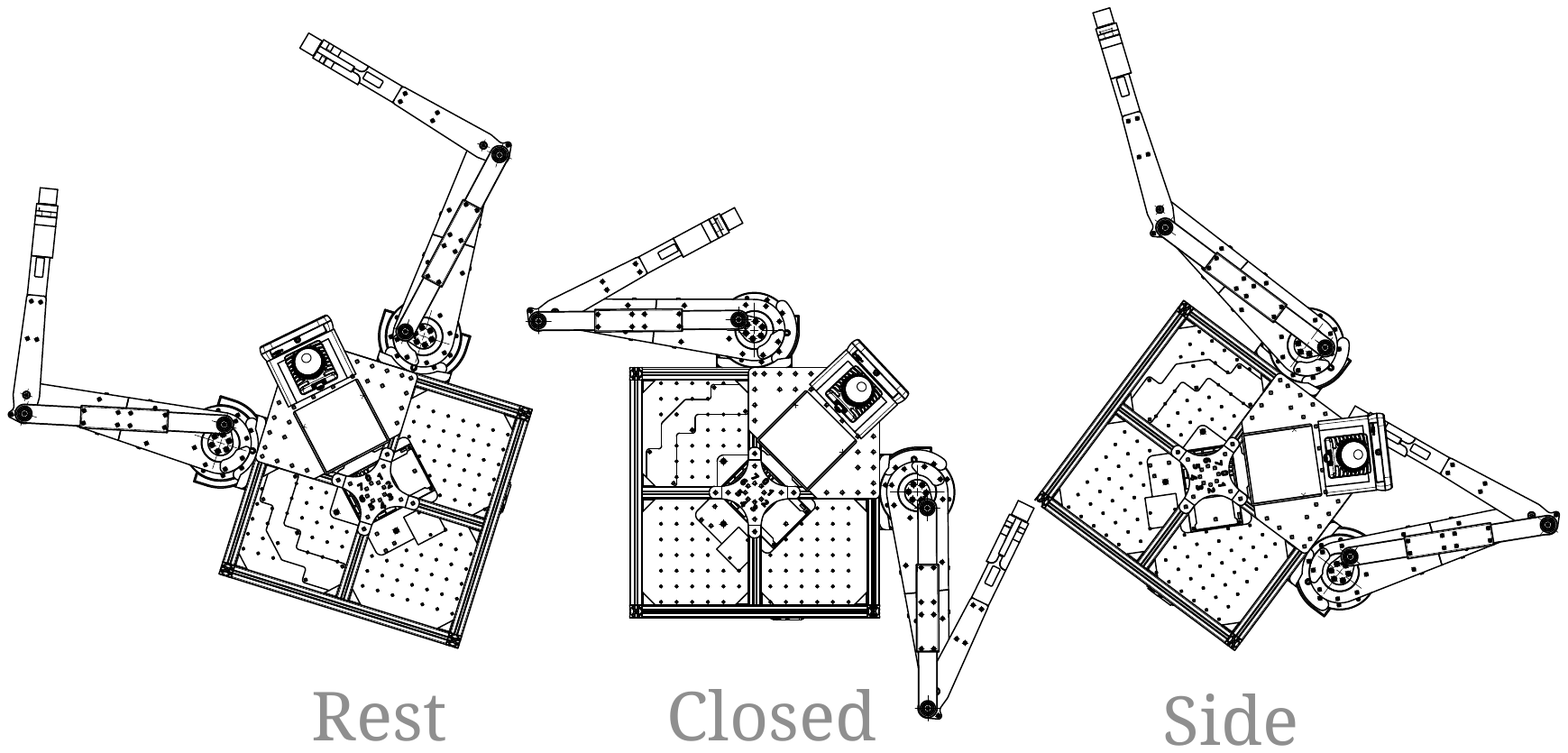}
    \caption{\textbf{\levion arm configurations.} Three static arm poses used throughout the experiments to vary the platform's effective moment of inertia: \emph{Rest} (arms extended outward at $90^{\circ}$, maximizing inertia); \emph{Closed} (arms tucked toward the body, minimizing inertia); and \emph{Side} (both arms extended to one side, producing an asymmetric mass distribution). The configurations are held fixed during each run and are not part of the controller's action space; they serve as a structured way to stress-test controller robustness across different inertial regimes.}
    \vspace{-1em}
    \label{fig:arms-mode}
\end{figure}

\paragraph{Task overview}
The four tasks are organized in increasing order of complexity. \textbf{Point-to-Pose Navigation} is the foundational planar maneuver, evaluated with both a classical LQR and a learned PPO controller across three actuator subsets (thrusters only, thrusters~+~reaction wheel, and all actuators), establishing baseline accuracy and a direct classical-vs-learned comparison across the three arm configurations of \cref{fig:arms-mode}. \textbf{Dynamic Disturbance Rejection} takes the same navigation task and adds a continuous, unobservable CoM disturbance driven by the \levion arms oscillating autonomously, isolating the learned policy's ability to reject structured, low-frequency inertia shifts that a rigid-body model cannot fully capture. \textbf{Momentum Damping / Stabilization} isolates the reaction wheel alone and evaluates a PPO policy's ability to damp arbitrary initial angular velocities across the three arm configurations, probing the wheel's real authority under varying effective inertia. \textbf{Force-Controlled Docking} is the most integration-intensive task: an ArUco marker detected by the onboard RGB-D camera provides a live 6-DoF goal to the learned point-to-pose controller, and contact with a fixed wall is managed through joint-space impedance control on the \levion arms, with contact onset detected by the wrist F/T sensors — exercising the full sensing and actuation stack in a single closed-loop pipeline.

\subsection{Tasks}

\subsubsection{Point-to-Pose Navigation}
\label{sec:exp:pose-nav}

In this task, \pingu must navigate from a random initial pose to a fixed target pose $(x^*, y^*, \psi^*)$. This is the platform's foundational maneuver and the first task on which both classical and learned controllers are evaluated side by side: every other task in this evaluation either builds directly on it (point-to-pose under disturbance, vision-guided docking) or reuses its reward structure. To highlight the system's reconfigurability, the task is executed using three actuator subsets: \emph{thrusters only}, \emph{thrusters and reaction wheel}, and \emph{all actuators} (including arms). For the two reduced subsets, the \levion arms are held static at one of the three fixed poses introduced in \cref{fig:arms-mode} (\emph{Side}, \emph{Rest}, \emph{Closed}), so that the effective inertia is varied independently of the controller. A classical LQR controller (with integral action over position and heading) and a learned PPO controller are both deployed on hardware for the \emph{thrusters only} and \emph{thrusters~+~reaction wheel} subsets; PPO alone is evaluated on the \emph{all-actuators} configuration since the arm joints are part of its action space. Performance is quantified by the final position error $e_p$ [m] and heading error $e_o$ [rad]. Results across all configurations are reported in \cref{tab:results_pose_nav}.

\begin{figure}[t]
    \centering
    \includegraphics[width=\linewidth]{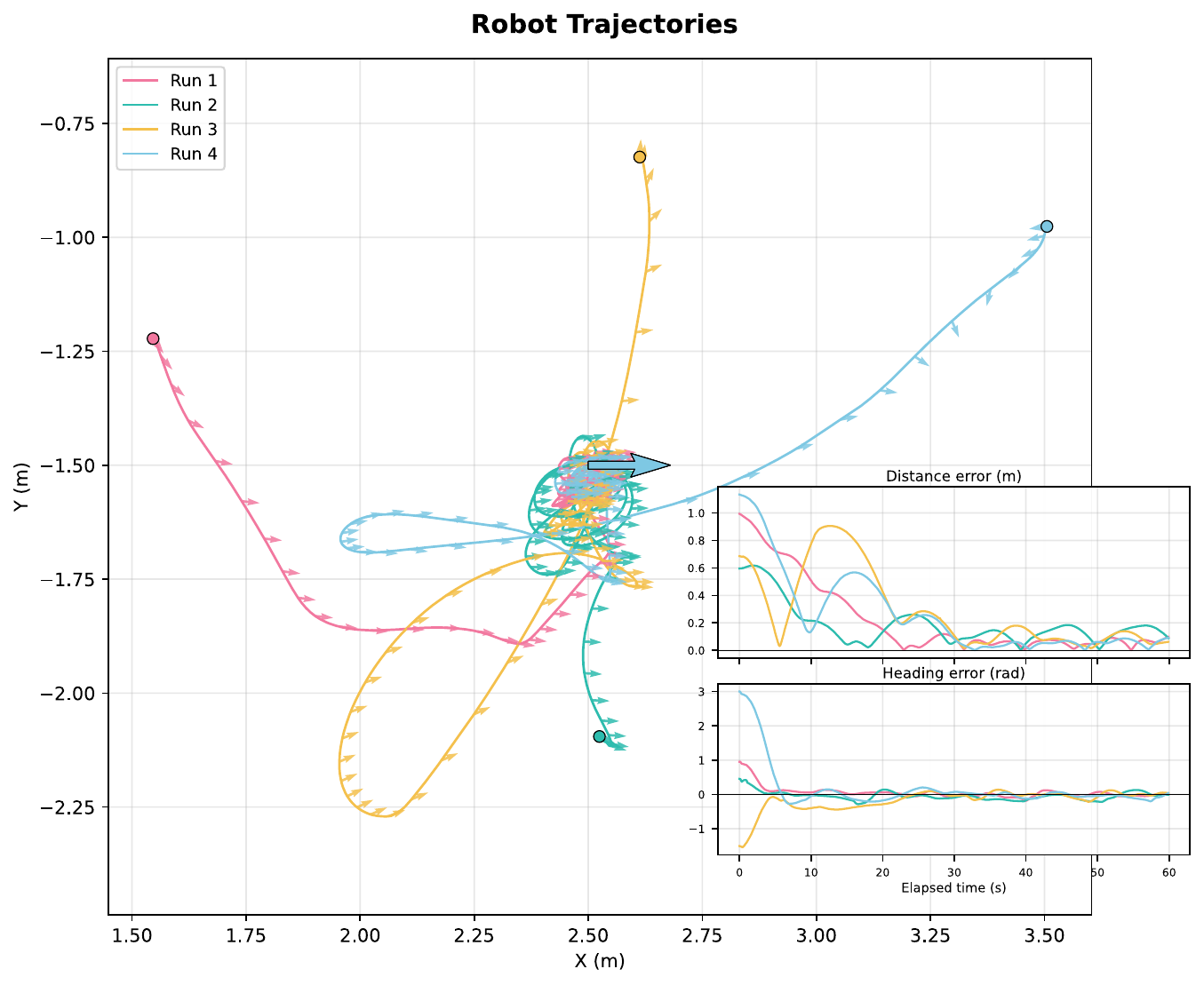}
    \caption{\textbf{Point-to-pose navigation.} Left: hardware trajectories of \pingu navigating from four random initial poses (colored dots) to a fixed target pose and heading (arrow), with arrowheads marking direction of travel. Right: corresponding distance error (top) and heading error (bottom) over time for the same four runs, both converging to near zero well within the 60\,s trial.}
    \label{fig:point-pose-nav}
\end{figure}

Because this reward is reused, unmodified, by the disturbance-rejection task and underlies the navigation component of the docking task, we define it in full here. For the learning controllers, the reward function $r_t^{\text{pose-nav}}$ is $$r_t^{\text{pose-nav}} = r_t^{\text{pose}} + r_t^{v,\omega} + r_t^{\text{bnd}}$$ where $r_t^{\text{pose}}$ rewards the robot for reaching the target position and heading simultaneously, $r_t^{v,\omega}$ penalizes the robot for exceeding a desired range of linear and angular velocities, and $r_t^{\text{bnd}}$ penalizes the robot's proximity to a defined maximum distance boundary, keeping it within the workspace. The full reward expression and weights are given in \cref{table:reward_functions}.

\begin{table}[h!]
    \centering
    \setlength{\tabcolsep}{7pt}
    \renewcommand{\arraystretch}{1.2}
    \resizebox{0.48\textwidth}{!}{
        \begin{tabular}{l *{2}{c} }
        \toprule
        \multirow{3}{*}{\textbf{Methods}}
        & \multicolumn{2}{c}{\textit{Point-to-Pose Navigation}} \\
        & $e_p [m] \downarrow$ & $e_o [rad] \downarrow$ \\
        \midrule
        \ourrow \multicolumn{2}{l}{\textbf{(a) All actuators}} & \\
        \cdashline{1-3}\noalign{\vskip 0.6mm}
            Sim
                & 0.0022 \ci{0.0015} & 0.0044 \ci{0.0033} \\
            Sim + DR
                & 0.0124  \ci{0.0191} & 0.0129  \ci{0.0124} \\
            PPO (Real)
                & 0.0081 \ci{0.0006} & 0.0257 \ci{0.0146} \\
        \midrule

        \ourrow \multicolumn{2}{l}{\textbf{(b) Thrusters and reaction wheel}} & \\
        \cdashline{1-3}\noalign{\vskip 0.6mm}
        \ourrow \multicolumn{3}{c}{\textit{Rest}} \\
            Sim
                & 0.0220  \ci{0.0217} & 0.0047  \ci{0.0036} \\
            Sim + DR
                &  0.0097  \ci{0.0098} & 0.0069  \ci{0.0059} \\
            PPO (Real)
                & 0.0482 \ci{0.0271} & 0.0279 \ci{0.0204} \\
            LQR (Real)
                & 0.1088 \ci{0.0837} & 0.0630 \ci{0.0954} \\
        \ourrow \multicolumn{3}{c}{\textit{Closed}} \\
            Sim
                & 0.0045  \ci{0.0055} & 0.0040  \ci{0.0033} \\
            Sim + DR
                & 0.0145  \ci{0.0111} & 0.0067  \ci{0.0072} \\
            PPO (Real)
                & 0.0679 \ci{0.0500} & 0.0598 \ci{0.0345} \\
            LQR (Real)
                & 0.1303 \ci{0.0763} & 0.1005 \ci{0.1179} \\
        \ourrow \multicolumn{3}{c}{\textit{Side}} \\
            Sim
                & 0.0137  \ci{0.0112} & 0.0047  \ci{0.0039} \\
            Sim + DR
                & 0.0083  \ci{0.0090} & 0.0081  \ci{0.0085} \\
            PPO (Real)
                & 0.0547 \ci{0.0422} & 0.0275 \ci{0.0154} \\
            LQR (Real)
                & 0.097 \ci{0.0574} & 0.0306 \ci{0.0111} \\
        \midrule

        \ourrow \multicolumn{2}{l}{\textbf{(c) Thrusters only}} & \\
        \cdashline{1-3}\noalign{\vskip 0.6mm}
        \ourrow \multicolumn{3}{c}{\textit{Rest}} \\
            Sim
                & 0.0057 \ci{0.0089} & 0.0046 \ci{0.0032} \\
            Sim + DR
                & 0.0056  \ci{0.0065} & 0.0105  \ci{0.0081} \\
            PPO (Real)
                & 0.0745 \ci{0.0396} & 0.0219 \ci{0.0028} \\
            LQR (Real)
                & 0.0843 \ci{0.0364} & 0.0461 \ci{0.0667} \\
        \ourrow \multicolumn{3}{c}{\textit{Closed}} \\
            Sim
                & 0.0099  \ci{0.0111} & 0.0055  \ci{0.0041} \\
            Sim + DR
                & 0.0075  \ci{0.0088} & 0.0062  \ci{0.0051} \\
            PPO (Real)
                & 0.0527 \ci{0.0192} & 0.0414 \ci{0.0363} \\
            LQR (Real)
                & 0.1696 \ci{0.0842} & 0.0307 \ci{0.0158} \\
        \ourrow \multicolumn{3}{c}{\textit{Side}} \\
            Sim
                & 0.0064 \ci{0.0106} & 0.0024 \ci{0.0026} \\
            Sim + DR
                & 0.0069 \ci{0.0087} & 0.0087 \ci{0.0082} \\
            PPO (Real)
                & 0.0241 \ci{0.0160} & 0.0335 \ci{0.0155} \\
            LQR (Real)
                & 0.0830 \ci{0.0248} & 0.0229 \ci{0.0234} \\
        \bottomrule
        \end{tabular}
    }
    \caption{\textbf{Point-to-pose navigation.} Final position error $e_p$ and heading error $e_o$ comparing the learned PPO controller (Sim, Sim+DR, and Real) against the classical LQR controller (Real only), across three actuator subsets and three static \levion arm poses (Side, Rest, Closed). The all-actuators configuration has no LQR baseline as the arms are part of the PPO action space.}
        \label{tab:results_pose_nav}
    \vspace{-1em}
\end{table}

We demonstrate the \pingu platform's ability to host both model-based and learned controllers through the same actuator-abstraction layer without any hardware modification by successfully deploying both PPO \cite{schulman2017proximalpolicyoptimizationalgorithms} and LQR\cite{kalman1960contributions}, and showing that converge to the target pose. The PPO controller consistently outperforms LQR on position accuracy across all actuator subsets and arm configurations: in the \emph{thrusters-only} configuration, PPO reduces $e_p$ by $55$--$69\%$ relative to LQR (e.g., $0.024$\,m vs.\ $0.083$\,m for Side, $0.053$\,m vs.\ $0.170$\,m for Closed); in the \emph{thrusters~+~reaction wheel} configuration, PPO similarly halves the position error (Rest: $0.048$\,m vs.\ $0.109$\,m; Closed: $0.068$\,m vs.\ $0.130$\,m). Heading errors are more comparable between the two controllers, with LQR occasionally matching PPO on that metric alone. The \emph{all-actuators} configuration, available to PPO only, achieves the lowest real-world position error overall ($e_p=0.0081$\,m), reflecting the additional control authority the arms provide when included in the action space. The sim-to-real gap for PPO is consistent across the three static arm poses within a given actuator subset, suggesting it is driven by unmodeled thruster and air-bearing dynamics rather than by the specific arm configuration.

\subsubsection{Dynamic Disturbance Rejection (CoM Shifting)}

\begin{figure}[t]
    \centering
    \includegraphics[width=\linewidth]{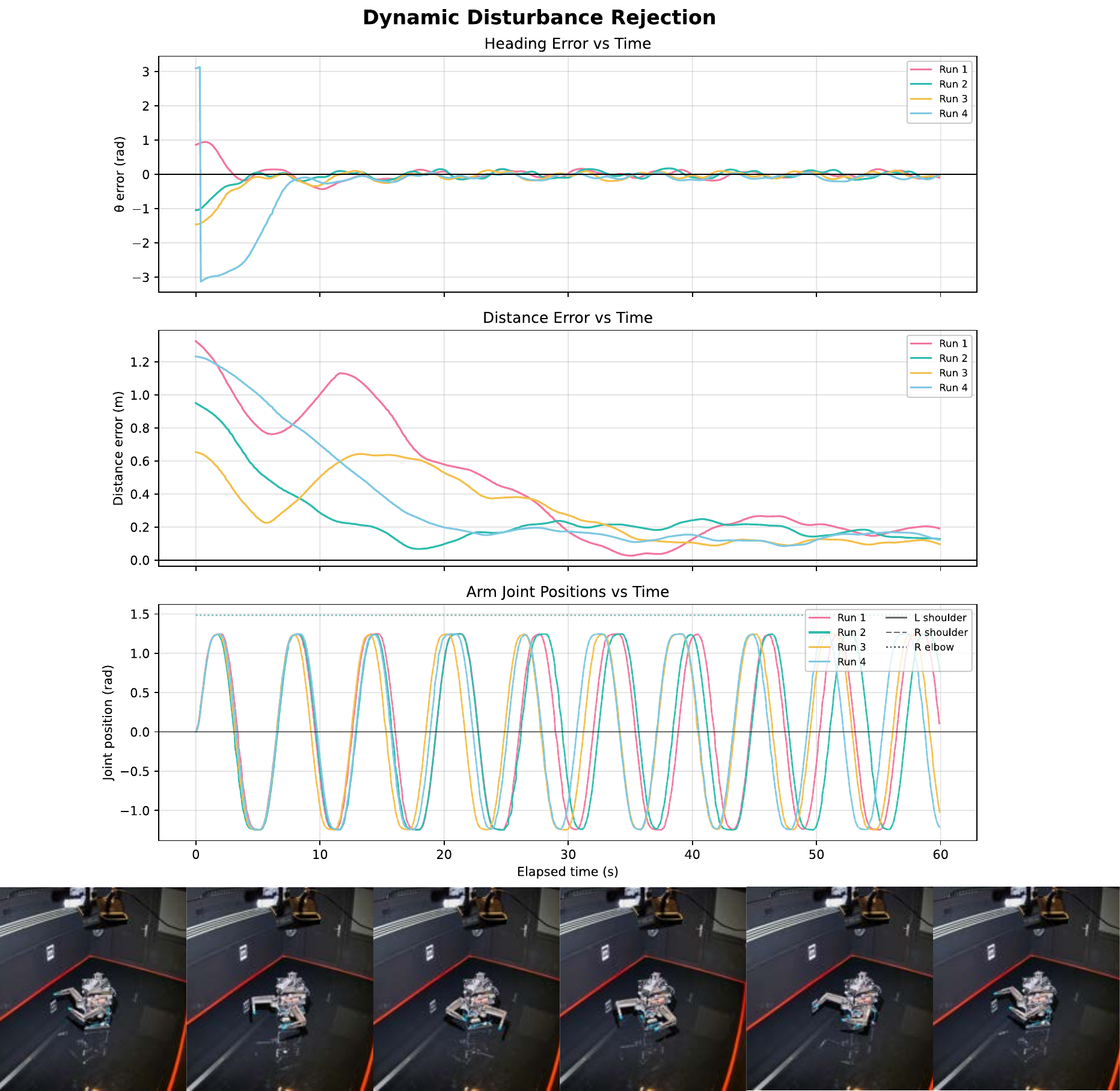}
    \caption{\textbf{Dynamic disturbance rejection.} Top: heading error $\theta$ over time for four representative hardware runs, converging toward zero despite the continuous CoM disturbance. Middle: corresponding distance error to the target pose, converging with bounded residual oscillations driven by the disturbance. Bottom: commanded shoulder/elbow joint positions for the same four runs, showing the disturbance generator's continuous $\pm0.8$\,rad oscillation throughout each 60\,s trial. Frame sequence: a representative run, with the platform navigating to the target pose while its arms swing autonomously.}
    \vspace{-1em}
    \label{fig:dynamic-disturbance-rejection}
\end{figure}

In this task, \pingu must navigate to a target pose $(x^*, y^*, \psi^*)$ while its two \levion arms introduce continuous, uncontrolled disturbances to the system's CoM. The arms oscillate autonomously between $\pm0.8$\,rad at the shoulder joints through a four-state finite state machine (hold right $\to$ swing left $\to$ hold left $\to$ swing right), with each transition duration sampled uniformly from $[0.1, 0.3]$\,s and cosine-interpolated to produce smooth, time-varying CoM displacements. Because the arms are driven by a separate position controller and are not part of the policy's observation space, their motion constitutes an exogenous disturbance that the agent must reject implicitly through its thruster and reaction-wheel commands alone. The task otherwise follows the same reward function as the Point-to-Pose Navigation task. It specifically evaluates the agent's ability to reject a persistent, structured CoM perturbation without direct observability of its source, probing the robustness of the learned momentum-management strategy.

\begin{table}[h!]
    \centering
    \setlength{\tabcolsep}{7pt}
    \renewcommand{\arraystretch}{1.2}
    \resizebox{0.49\textwidth}{!}{
        \begin{tabular}{l *{2}{c} }
        \toprule
        \multirow{3}{*}{\textbf{Methods}}
        & \multicolumn{2}{c}{\textit{Dynamic Disturbance Rejection}} \\
        & $e_p [m] \downarrow$ & $e_o [rad] \downarrow$ \\
        \midrule
        Sim
            & 0.0070 \ci{0.0054} &  0.0074 \ci{0.0108} \\
        Sim + DR
            & 0.0156 \ci{0.0161} & 0.0340 \ci{0.0574} \\
        Real
            & 0.0889 \ci{0.0405} & 0.0346 \ci{0.0179} \\
        \bottomrule
        \end{tabular}
    }
    \caption{\textbf{Dynamic disturbance rejection.} Final position error $e_p$ and heading error $e_o$ for the PPO policy navigating to a fixed target pose under continuous, unobserved CoM disturbance from arm motion, evaluated in simulation, in simulation with domain randomization (DR), and on hardware.}
        \label{tab:results_dynamic_disturbance}
    \vspace{-1em}
\end{table}

The heading error degrades only mildly from simulation to hardware: domain randomization alone raises $e_o$ from $0.0074$\,rad (Sim) to $0.0340$\,rad (Sim~+~DR), and the real-hardware result ($e_o=0.0346$\,rad) is nearly indistinguishable from the Sim~+~DR value, indicating that domain randomization closes most of the heading sim-to-real gap. Position tracking is harder to transfer: the real position error ($e_p=0.0889$\,m) is more than five times larger than the Sim~+~DR value ($e_p=0.0156$\,m), a substantially larger jump than for heading. Because the policy never observes the arms directly, it must infer and reject the resulting CoM-induced positional drift purely from the platform's own pose feedback, so residual modeling mismatches in the real thruster and air-bearing dynamics are amplified by the disturbance's persistent, never-settling nature. Despite this gap, the policy maintains stable closed-loop tracking throughout each 60\,s trial, with the bounded distance-error oscillations visible in \cref{fig:dynamic-disturbance-rejection} confirming that it successfully rejects a persistent, unobserved CoM perturbation without diverging.

\subsubsection{Momentum Damping / Stabilization}

\begin{figure}[h]
    \centering
    \includegraphics[width=\linewidth]{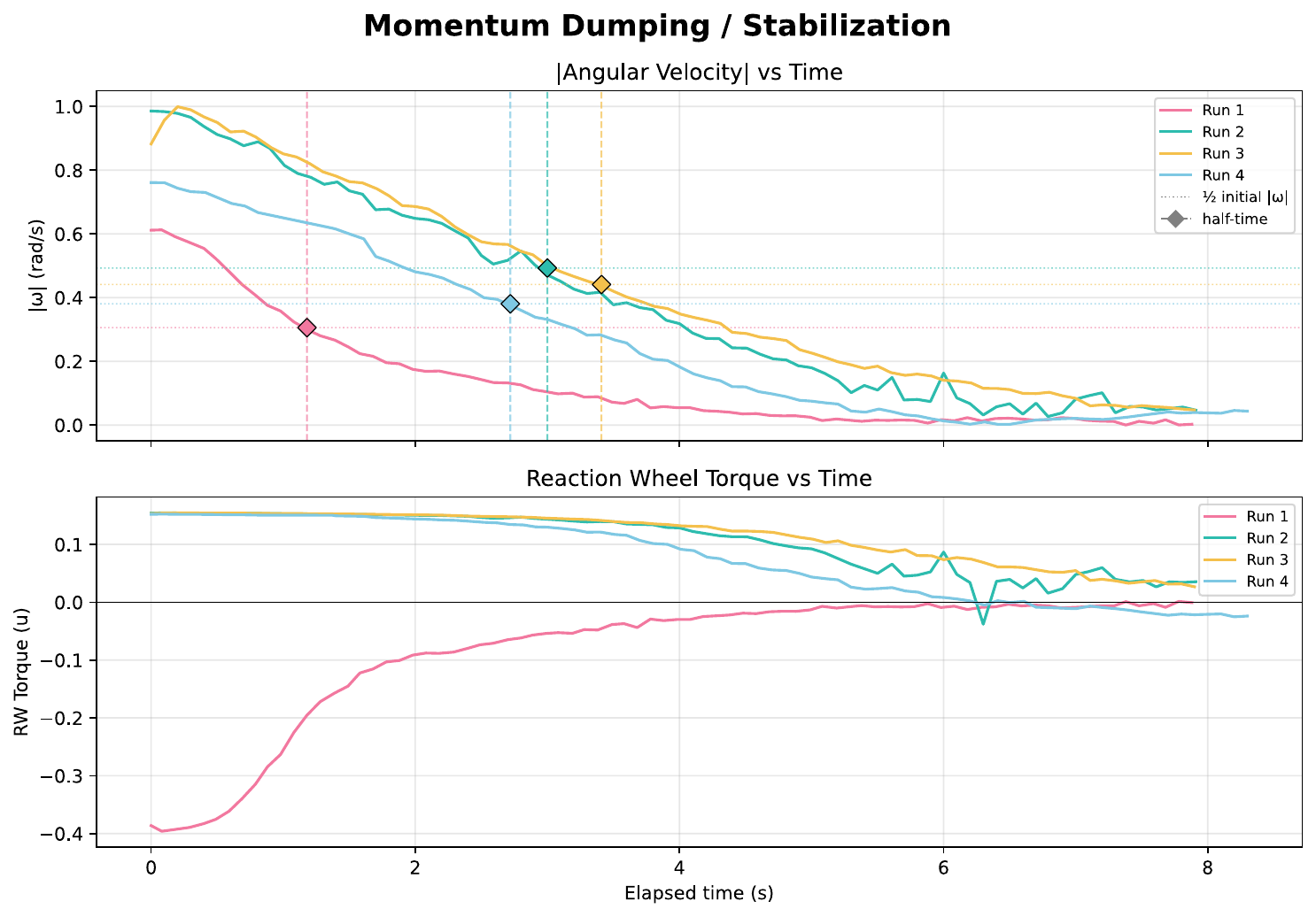}
    \caption{\textbf{Momentum damping in the \emph{Rest} arm configuration.} Top: magnitude of the platform's angular velocity over time across four representative hardware stabilization runs; dotted lines mark half the initial angular velocity and diamond markers indicate each run's half-time $t_{1/2}$. Bottom: corresponding commanded reaction-wheel torque for the same four runs, showing the PPO policy applying progressively less braking torque as the angular velocity decays toward zero.}
    \vspace{-1em}
    \label{fig:stabilization}
\end{figure}

In this task, \pingu is initialized with a random angular velocity drawn uniformly from $[\pm0.1,\,\pm0.5]$\,rad/s and must bring its rotation to rest using the reaction wheel alone, with no thruster actuation. The experiment probes the reaction wheel's angular-momentum absorption capacity across three \levion arm configurations that vary the platform's effective moment of inertia (\cref{fig:arms-mode}): \emph{Side} (producing an asymmetric mass distribution); \emph{Rest} (extended radially outward and maximizing effective inertia); and \emph{Closed} (minimizing effective inertia). A PPO policy is trained to command the reaction wheel torque as a function of the current angular velocity; \cref{fig:stabilization} illustrates the resulting angular-velocity decay and commanded torque for four representative runs in the \emph{Rest} configuration. Performance across all three configurations is quantified by the residual angular velocity $\phi$ [rad/s] and the half-time $t_{1/2}$ [s], the time for the angular velocity to reach half its initial magnitude, as reported in \cref{tab:results_stabilization}.

\begin{table}[h!]
    \centering
    \setlength{\tabcolsep}{7pt}
    \renewcommand{\arraystretch}{1.2}
    \resizebox{0.49\textwidth}{!}{
        \begin{tabular}{l *{3}{c} }
        \toprule
        \multirow{3}{*}{\textbf{Methods}}
        & \multicolumn{2}{c}{\textit{Stabilization}} \\
        & $\phi [rad/s] \downarrow$ & $t_{\text{half}} [s] \downarrow$ \\
        \midrule
        \ourrow \multicolumn{2}{l}{\textbf{(b) Rest}} & \\
        \cdashline{1-3}\noalign{\vskip 0.6mm}
        Sim
            & 0.0467  \ci{0.0153} & 3.7975  \ci{1.6068} \\
        Real
            & 0.0228 \ci{0.0229} & 6.377 \ci{4.941} \\
            
        \ourrow \multicolumn{2}{l}{\textbf{(c) Closed}} & \\
        \cdashline{1-3}\noalign{\vskip 0.6mm}
        Sim
            & 0.0432  \ci{0.0150} & 3.8773  \ci{1.6457} \\
        Real
            & 0.0534 \ci{0.0926} & 6.580 \ci{5.046} \\
        \ourrow \multicolumn{2}{l}{\textbf{(a) Side}} & \\
        \cdashline{1-3}\noalign{\vskip 0.6mm}
        Sim
            & 0.0429  \ci{0.0148} & 4.0773  \ci{4.7499} \\
        Real
            & 0.0594 \ci{0.0304} & 4.845 \ci{1.873} \\
        \bottomrule
        \end{tabular}
    }
    \caption{\textbf{Momentum damping / stabilization.} Residual angular velocity $\phi$ and half-time $t_{1/2}$ of the reaction-wheel-only PPO stabilization policy, evaluated in simulation and on hardware across three \levion arm configurations (Side, Rest, Closed) spanning the platform's range of effective moment of inertia.}
    \label{tab:results_stabilization}
    \vspace{-1em}
\end{table}

In simulation, the policy stabilizes the platform at a comparable rate and residual velocity across all three arm configurations ($t_{1/2} \approx 3.8$--$4.1$\,s, $\phi \approx 0.043$--$0.047$\,rad/s), indicating low sensitivity to the arm-induced inertia changes considered here. On hardware, the policy decays the angular velocity fastest in the \emph{Side} configuration ($t_{1/2}=4.845$\,s), as also illustrated by the representative \emph{Rest}-configuration runs in \cref{fig:stabilization}, but settles to the largest residual velocity of the three ($\phi=0.0594$\,rad/s); conversely, \emph{Rest} decays more slowly ($t_{1/2}=6.377$\,s) yet reaches the lowest residual velocity overall ($\phi=0.0228$\,rad/s), while \emph{Closed} performs worst on both metrics ($t_{1/2}=6.580$\,s, $\phi=0.0534$\,rad/s). This rank reversal between fastest initial decay and lowest long-term residual velocity, absent from the simulated results, indicates that the reaction wheel's real disturbance-rejection authority interacts with the platform's effective inertia distribution in ways not fully captured by the simulated dynamics, particularly in the low-velocity tail of the response.

\subsubsection{Force-Controlled Docking}

\begin{figure}[h]
    \centering
    \includegraphics[width=\linewidth]{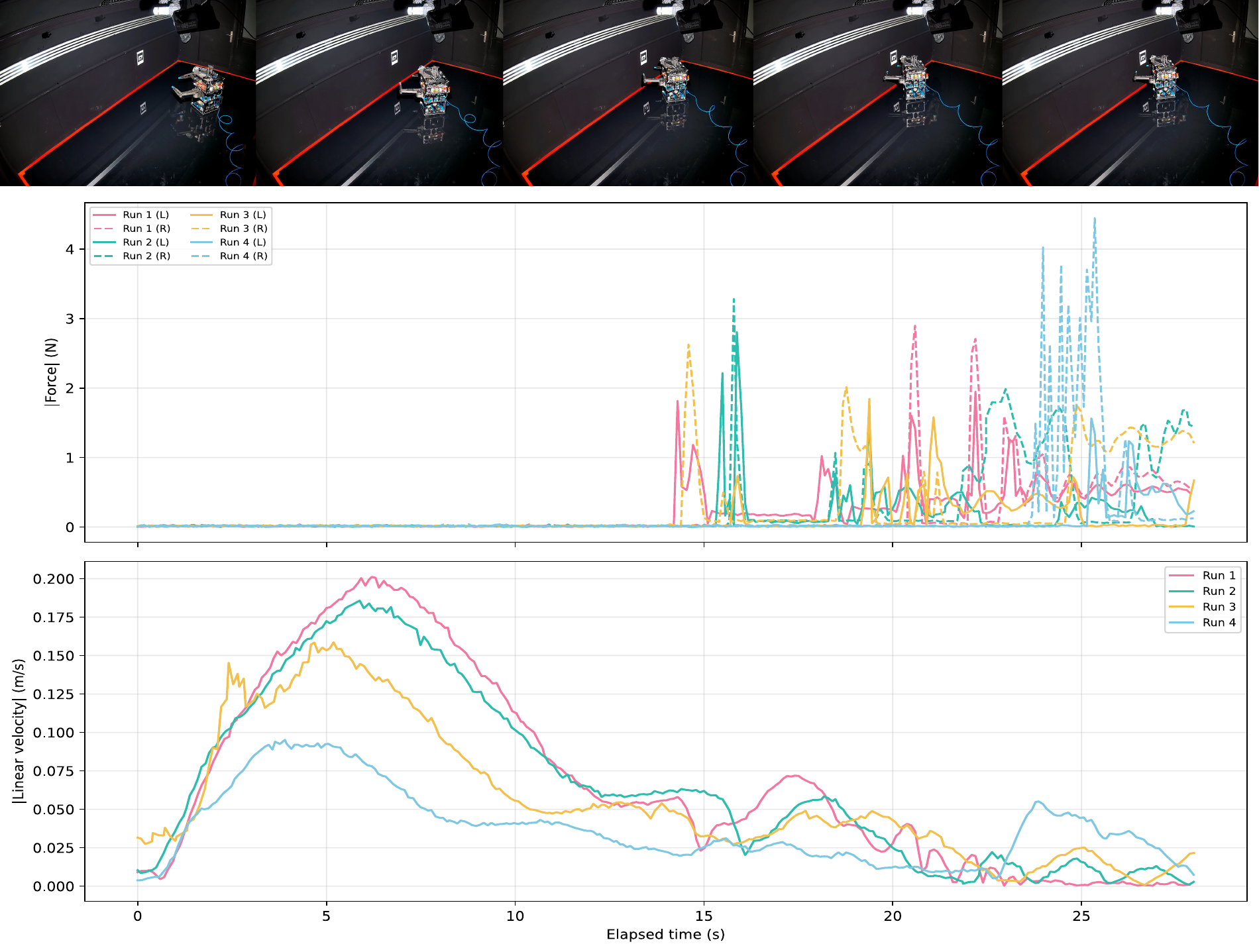}
    \caption{\textbf{Force-controlled docking.} Top: frame sequence of a representative run, with the platform's trajectory approaching the wall-mounted ArUco target. Middle: magnitude of the F/T-sensed contact force at the left (solid) and right (dashed) wrist over four repeated runs, showing intermittent contact spikes once the arms reach the wall. Bottom: magnitude of the platform's linear velocity for the same four runs, peaking mid-approach and decaying to near zero as the arms' compliant contact dissipates the platform's residual kinetic energy.}
    \vspace{-1em}
    \label{fig:wall_docking}
\end{figure}

This task showcases the integration of \pingu's full sensing and control stack into a single pipeline, chaining vision-based pose estimation, learned navigation control, and contact-triggered compliant control across three of the platform's actuation modalities. The Intel RealSense D455 RGB camera on the sensor payload detects an ArUco fiducial marker affixed to a fixed wall and estimates its 6-DoF pose relative to the platform in real time; this pose is fed as the goal to the same learned point-to-pose controller evaluated above, which drives the floating platform toward the wall. As the \levion arms make contact with the wall, the Leptrino six-axis F/T sensor at each wrist registers the contact event. Rather than actively braking with the thrusters or reaction wheel, \pingu lets its own inertia carry the residual approach velocity into the wall, and the resulting contact is damped entirely at the arm level by a joint-space impedance controller running in the \levion arms' ros2\_control loop: each joint tracks a fixed position setpoint through a virtual spring-damper law with fixed $K_p$ and $K_d$ gains, so that the arm yields compliantly on contact and dissipates the platform's residual kinetic energy without letting the wall's reaction force push the floating base back away from the docking surface. \cref{fig:wall_docking} reports results over multiple docking runs, showing the F/T force measured at the left and right wrists together with the platform's velocity at the end of each contact.

\section{Discussion}
\label{sec:discussion}

The experiments of \cref{sec:experiments} were designed less to crown a single best controller than to demonstrate that \pingu can host a heterogeneous set of controllers and tasks on one chassis. Several lessons emerge from running them.

\paragraph{Classical and learned control share the same hardware, with complementary strengths}
The headline systems result is that a hand-tuned LQR and an end-to-end PPO policy both deploy on \pingu through the identical actuator-abstraction layer, on identical tasks, with no change to the low-level stack (\cref{sec:exp:pose-nav}). Within that common envelope the two controllers exhibit a clear and consistent trade-off: the learned policy roughly halves the final position error of LQR across every actuator subset and arm configuration, while the two are far closer on heading, where LQR occasionally matches PPO. The advantage widens when the arms are admitted into the policy's action space --- the all-actuators configuration attains the lowest position error in the study ($e_p = 0.0081$\,m), confirming that the additional control authority of the arms is usable by a learned controller in a way an LQR built on a fixed rigid-body model cannot exploit. This is precisely the kind of like-for-like comparison \pingu is built to enable, and it is, to our knowledge, the first reported across the full thruster\,+\,reaction-wheel\,+\,arm envelope on real hardware.

\paragraph{Real attitude authority depends on effective inertia in ways simulation under-predicts}
The reaction-wheel stabilization task (\cref{tab:results_stabilization}) is the clearest example of a phenomenon that only the hardware reveals. In simulation the policy damps the platform at near-identical rates across all three arm configurations; on hardware a rank reversal appears between fastest initial decay (\emph{Side}, $t_{1/2}=4.85$\,s) and lowest long-term residual velocity (\emph{Rest}, $\phi=0.0228$\,rad/s), with the minimum-inertia \emph{Closed} pose worst on both metrics. The reaction wheel's true disturbance-rejection authority therefore interacts with the platform's mass distribution and with friction in the low-velocity tail in a manner the simulated dynamics do not fully capture. That this surfaces only on hardware underscores the value of a platform on which effective inertia can be varied as a controlled, repeatable experimental variable rather than a modeling assumption.

\paragraph{Where the sim-to-real gap lives}
Across tasks the gap is strongly anisotropic. Domain randomization closes almost all of the \emph{heading} gap --- in disturbance rejection the real heading error ($0.0346$\,rad) is nearly indistinguishable from the Sim\,+\,DR value ($0.0340$\,rad) --- whereas \emph{position} tracking transfers far less cleanly, with the real error exceeding Sim\,+\,DR by more than fivefold under a persistent CoM disturbance. Two observations localize the cause. First, the gap is consistent across the three static arm poses within a given actuator subset, which points away from the arm-induced inertia change and toward unmodeled thruster and air-bearing dynamics as the dominant residual. The shared-manifold pressure coupling characterized in \cref{fig:thruster-charact}, whereby simultaneous valve firing depresses per-thruster force, is a prime suspect: it directly corrupts the translational impulse budget while leaving differential yaw commands comparatively intact. Second, the low-rate 10 Hz policy-inference loop, set against 100 Hz state feedback and a 200 Hz joint loop, bounds how quickly translational drift can be corrected once it accumulates. Both are concrete, characterizable targets --- manifold-aware thrust allocation and a higher-rate inference path --- rather than fundamental limits.

\paragraph{Reconfigurable inertia as a first-class research instrument}
The \levion arms serve two distinct roles in these experiments, and both proved essential. Held static in the three poses of \cref{fig:arms-mode}, they let us sweep effective inertia without touching the hardware; driven autonomously, they generate a continuous, unobserved CoM disturbance that the navigation policy must reject blind (\cref{fig:dynamic-disturbance-rejection}). The policy maintains bounded, non-divergent tracking throughout each trial despite never observing the arms --- evidence that learned controllers can absorb the structured, hard-to-model inertia variation that classical model-based design handles least gracefully. This is the central capability distinguishing \pingu from thruster-and-wheel-only emulators, and it is what makes the platform a substrate for the open problem of sim-to-real RL under reconfigurable inertia.

\paragraph{Integration and maintainability}
Finally, the experiments are themselves evidence for the system-level design choices. The two-workspace ROS\,2 architecture (\cref{fig:software-architecture}) meant that swapping LQR for PPO, or a fixed marker goal for a live RGB-D detection in the docking task, required relaunching only controller-layer nodes; the hardware-abstraction and sensor layers were untouched across all four tasks. The two-domain power split kept high-current arm and wheel transients off the control rails over long experiment campaigns. These are not headline scientific results, but they determine whether a platform survives contact with real research use --- and they are the properties we most hope the open-source release transfers to other groups.

\section{Conclusion}
\label{sec:conclusion}

We presented \pingu, a reconfigurable planar microgravity platform that co-integrates eight cold-gas thrusters, a reaction wheel, and two force/torque-sensed \levion arms with interchangeable end-effectors that extends the open-source ATMOS testbed, under a unified ROS\,2 stack that exposes every actuator through a common abstraction layer. We characterized each subsystem, described the control and simulation infrastructure, and validated the platform on four tasks --- point-to-pose navigation, dynamic disturbance rejection under arm-induced CoM shifts, reaction-wheel stabilization, and force-controlled docking --- deploying classical and learned controllers on the same hardware through the same interface. Taken together, the results support the paper's central claim: \pingu is less a single result than an \emph{enabler}, a common substrate on which the spacecraft-GNC and reinforcement-learning communities can pose and answer questions that no existing planar testbed supports end-to-end.

That framing is clearest when set against the representative use cases of \cref{fig:pingu-overview}. Of the five envisioned, three --- docking, point navigation, and momentum stabilization --- are demonstrated here; the remaining two define our immediate roadmap.

\paragraph{Dual-arm capture and detumbling}
The \emph{target-catching} use case is the natural next step for the arms: coordinated dual-arm capture and detumbling of a free, tumbling target, with the wrist F/T sensors closing the compliant-impact loop that contact-rich capture demands~\cite{nagaoka2018dualarm,hovell2022capture,Santaguida2023}. The ceiling- and wall-mounted UR10 manipulators already modeled in the digital twin (\cref{fig:digital-twin}) provide a repeatable, programmable moving target for exactly this scenario, letting capture policies be trained in simulation and transferred to a controlled physical rehearsal.

\paragraph{Magnetic wall locomotion}
The \emph{wall-locomotion} use case exploits \pingu's standardized wrist coupling: fitting magnetic end-effectors lets the arms ``walk'' the platform along a ferromagnetic structure, a propellant-free locomotion and station-keeping mode for inspection and traversal of large structures. This recasts the arms from manipulators into a locomotion subsystem and stresses the base--arm coupling in a regime --- sustained contact under intermittent anchoring --- that none of our current tasks exercises.

\paragraph{Richer perception in the loop}
The docking experiment relied on a single ArUco marker and the RGB-D camera; the platform's event camera, LiDAR, and global-shutter RGB sensor remain outside the control loop. Bringing them in --- for markerless, high-rate relative pose estimation of non-cooperative targets --- is the perception counterpart to the capture and locomotion goals above, and the multi-modal sensor board was provisioned with exactly this progression in mind.

We release the full hardware design, firmware, ROS\,2 packages, simulation environments, and experimental logs to lower the barrier to entry for learning-based spacecraft GNC research, in the hope that \pingu serves the community as a shared, extensible foundation for the contact-rich, inertia-coupled proximity-operations problems that remain open.

\section*{Acknowledgments}
The authors would like to thank the ATMOS team for their invaluable help and support throughout this work.

\bibliographystyle{unsrtnat}
\bibliography{references}

@inproceedings{Kwok-Choon2018Design,
  title     = {Design, Fabrication, and Preliminary Testing of Air-Bearing Test Vehicles for the Study of Autonomous Satellite Maneuvers},
  author    = {Kwok-Choon, Stephen and Buchala, Kathleen and Blackwell, Briana and Lopresti, Steven and Wilde, Markus and Go, Thein Htay},
  booktitle = {Proceedings of the 31st Florida Conference on Recent Advances in Robotics (FCRAR)},
  year      = {2018},
  address   = {Orlando, FL, USA},
  month     = {May},
  pages     = {25},
  note      = {The paper number in the proceedings is 25.}
}

@article{Schwartz2003Historical,
  author  = {Schwartz, Jana L. and Peck, Mason A. and Hall, Christopher D.},
  title   = {Historical Review of Air-Bearing Spacecraft Simulators},
  journal = {Journal of Guidance, Control, and Dynamics},
  volume  = {26},
  number  = {4},
  pages   = {513--522},
  year    = {2003},
  doi     = {10.2514/2.5085}
}

@article{rybus2016planar,
  author  = {Rybus, Tomasz and Seweryn, Karol},
  title   = {Planar Air-Bearing Microgravity Simulators: Review of Applications, Existing Solutions and Design Parameters},
  journal = {Acta Astronautica},
  volume  = {120},
  pages   = {239--259},
  year    = {2016}
}

@misc{roque2025opensourcemodularspacesystems,
      title={Towards Open-Source and Modular Space Systems with ATMOS}, 
      author={Pedro Roque and Sujet Phodapol and Elias Krantz and Jaeyoung Lim and Joris Verhagen and Frank J. Jiang and David Dörner and Huina Mao and Gunnar Tibert and Roland Siegwart and Ivan Stenius and Jana Tumova and Christer Fuglesang and Dimos V. Dimarogonas},
      year={2025},
      eprint={2501.16973},
      archivePrefix={arXiv},
      primaryClass={cs.RO},
      url={https://arxiv.org/abs/2501.16973}, 
}

@article{yalccin2023lightweight,
  title={Lightweight floating platform for ground-based emulation of on-orbit scenarios},
  author={Yal{\c{c}}{\i}n, Bar{\i}{\c{s}} Can and Martinez, Carol and Coloma, Sof{\'\i}a and Skrzypczyk, Ernest and Olivares-Mendez, Miguel A},
  journal={IEEE Access},
  volume={11},
  pages={94575--94588},
  year={2023},
  publisher={IEEE}
}

@misc{bredenbeck2022findingfollowingoptimaltrajectories,
      title={Finding and Following Optimal Trajectories for an Overactuated Floating Robotic Platform}, 
      author={Anton Bredenbeck and Shubham Vyas and Willem Suter and Martin Zwick and Dorit Borrmann and Miguel Olivares-Mendez and Andreas Nüchter},
      year={2022},
      eprint={2206.03993},
      archivePrefix={arXiv},
      primaryClass={cs.RO},
      url={https://arxiv.org/abs/2206.03993}, 
}

@article{banerjee2022floating,
  author  = {Banerjee, Avijit and Satpute, Sumeet G. and Kanellakis, Christoforos and Tevetzidis, Ilias and Haluska, Jakub and Bodin, Per and Nikolakopoulos, George},
  title   = {On the Design, Modeling and Experimental Verification of a Floating Satellite Platform},
  journal = {IEEE Robotics and Automation Letters},
  volume  = {7},
  number  = {2},
  pages   = {1364--1371},
  year    = {2022},
  doi     = {10.1109/LRA.2021.3140134}
}

@inproceedings{Nakka2018ASD,
  author    = {Nakka, Yashwanth Kumar and Foust, Rebecca C. and Lupu, Elena Sorina and Elliott, David B. and Crowell, Irene S. and Chung, Soon-Jo and Hadaegh, Fred Y.},
  title     = {A Six Degree-of-Freedom Spacecraft Dynamics Simulator for Formation Control Research},
  booktitle = {AAS/AIAA Astrodynamics Specialist Conference (AAS 18-476)},
  year      = {2018}
}

@inproceedings{Tsiotras2014ASTROS,
  author    = {Tsiotras, Panagiotis},
  title     = {{ASTROS}: A 5{D}o{F} Experimental Facility for Research in Space Proximity Operations},
  booktitle = {AAS Guidance, Navigation and Control Conference (AAS 14-114)},
  year      = {2014}
}

@inproceedings{Schlotterer2010Testbed,
  author    = {Schlotterer, Markus and Theil, Stephan},
  title     = {Testbed for On-Orbit Servicing and Formation Flying Dynamics Emulation},
  booktitle = {AIAA Guidance, Navigation, and Control Conference},
  year      = {2010},
  doi       = {10.2514/6.2010-8108}
}

@article{Eun2018Development,
  author  = {Eun, Youngho and Park, Sang-Young and Kim, Geuk-Nam},
  title   = {Development of a Hardware-in-the-Loop Testbed to Demonstrate Multiple Spacecraft Operations in Proximity},
  journal = {Acta Astronautica},
  volume  = {147},
  pages   = {48--58},
  year    = {2018}
}

@article{Huang2022Characterizing,
  author  = {Huang, Zheng and Zhang, Wei and Chen, Ti and Wen, Hao and Jin, Dongping},
  title   = {Characterizing an Air-Bearing Testbed for Simulating Spacecraft Dynamics and Control},
  journal = {Aerospace},
  volume  = {9},
  number  = {5},
  pages   = {246},
  year    = {2022}
}

@inproceedings{Kolvenbach2014Orbit,
  author    = {Kolvenbach, Hendrik and Wormnes, Kjetil},
  title     = {Recent Developments on {ORBIT}, a 3-{D}o{F} Free Floating Contact Dynamics Testbed},
  booktitle = {Proc. ESA Symp. Advanced Space Technologies in Robotics and Automation (ASTRA)},
  year      = {2015}
}

@inproceedings{Papadopoulos2015NTUA,
  author    = {Papadopoulos, Evangelos and Paraskevas, Iosif S. and Flessa, Thaleia and Nanos, Kostas and Rekleitis, Georgios and Kontolatis, Ioannis},
  title     = {The {NTUA} Space Robot Simulator: Design \& Results},
  booktitle = {Proc. ESA Symp. Advanced Space Technologies in Robotics and Automation (ASTRA)},
  year      = {2015}
}

@article{Ragan2024OnlineTree,
  author  = {Ragan, James and Riviere, Benjamin and Hadaegh, Fred Y. and Chung, Soon-Jo},
  title   = {Online Tree-Based Planning for Active Spacecraft Fault Estimation and Collision Avoidance},
  journal = {Science Robotics},
  year    = {2024}
}

@article{Sabatini2017Coordinated,
  author  = {Sabatini, Marco and Gasbarri, Paolo and Palmerini, Giovanni B.},
  title   = {Coordinated Control of a Space Manipulator Tested by Means of an Air Bearing Free Floating Platform},
  journal = {Acta Astronautica},
  volume  = {139},
  pages   = {296--305},
  year    = {2017}
}

@article{Santaguida2023,
  author  = {Santaguida, Lucas and Zhu, Zheng H.},
  title   = {Development of Air-Bearing Microgravity Testbed for Autonomous Spacecraft Rendezvous and Robotic Capture Control of a Free-Floating Target},
  journal = {Acta Astronautica},
  volume  = {203},
  pages   = {319--328},
  year    = {2023}
}

@inproceedings{bualat2018astrobee,
  author    = {Bualat, Maria G. and Smith, Trey and Smith, Ernest E. and Fong, Terrence and Wheeler, D. W.},
  title     = {{Astrobee}: A New Tool for {ISS} Operations},
  booktitle = {AIAA SpaceOps Conference},
  year      = {2018}
}

@inproceedings{smith2016astrobee,
  author    = {Smith, Trey and Barlow, Jonathan and Bualat, Maria and Fong, Terrence and Provencher, Christopher and Sanchez, Hugo and Smith, Ernest},
  title     = {{Astrobee}: A New Platform for Free-Flying Robotics on the {ISS}},
  booktitle = {Int. Symp. Artificial Intelligence, Robotics and Automation in Space (i-SAIRAS)},
  year      = {2016}
}

@inproceedings{Turchetti2024IVFFS,
  author    = {Turchetti, Federico and Ekal, Monica and Lii, Neal Y. and Roa, Maximo A.},
  title     = {Analysis of Intra-Vehicular Robotic Free-Flyers and Their Manipulation Capabilities},
  booktitle = {75th Int. Astronautical Congress (IAC), Milan},
  year      = {2024}
}

@inproceedings{ElHariry2024DRIFT,
  author    = {El-Hariry, Matteo and Richard, Antoine and Muralidharan, Vivek and Geist, Matthieu and Olivares-Mendez, Miguel},
  title     = {{DRIFT}: Deep Reinforcement Learning for Intelligent Floating Platforms Trajectories},
  booktitle = {arXiv:2310.04266v2},
  year      = {2024}
}

@article{Gaudet2020DRLLanding,
  author  = {Gaudet, Brian and Linares, Richard and Furfaro, Roberto},
  title   = {Deep Reinforcement Learning for Six Degree-of-Freedom Planetary Landing},
  journal = {Advances in Space Research},
  volume  = {65},
  number  = {7},
  pages   = {1723--1741},
  year    = {2020}
}

@inproceedings{Hovell2020DRL,
  author    = {Hovell, Kirk and Ulrich, Steve},
  title     = {On Deep Reinforcement Learning for Spacecraft Guidance},
  booktitle = {AIAA SciTech Forum},
  year      = {2020}
}

@article{Athauda2023RL,
  author  = {Athauda, Dinusha and Banerjee, Avijit and Satpute, Sumeet and Nikolakopoulos, George},
  title   = {Intelligent Motion Planning for Collision Free Autonomous Docking of Satellite Emulation Platform Using Reinforcement Learning},
  journal = {IFAC-PapersOnLine},
  volume  = {56},
  number  = {2},
  pages   = {3354--3359},
  year    = {2023}
}

@article{Cao2023RLDualArm,
  author  = {Cao, Yang and Wang, Shuang and others},
  title   = {Reinforcement Learning with Prior Policy Guidance for Motion Planning of Dual-Arm Free-Floating Space Robot},
  journal = {Aerospace Science and Technology},
  volume  = {136},
  pages   = {108098},
  year    = {2023}
}

@article{Umetani1989Resolved,
  author  = {Umetani, Yoji and Yoshida, Kazuya},
  title   = {Resolved Motion Rate Control of Space Manipulators with Generalized {J}acobian Matrix},
  journal = {IEEE Transactions on Robotics and Automation},
  volume  = {5},
  number  = {3},
  pages   = {303--314},
  year    = {1989}
}

@article{Papadopoulos1991Dynamics,
  author  = {Papadopoulos, Evangelos and Dubowsky, Steven},
  title   = {On the Nature of Control Algorithms for Free-Floating Space Manipulators},
  journal = {IEEE Transactions on Robotics and Automation},
  volume  = {7},
  number  = {6},
  pages   = {750--758},
  year    = {1991}
}

@inproceedings{Liao2024BerkeleyHumanoid,
  author    = {Liao, Qiayuan and Zhang, Bike and Huang, Xuanyu and Huang, Xiaoyu and Li, Zhongyu and Sreenath, Koushil},
  title     = {{Berkeley Humanoid}: A Research Platform for Learning-Based Control},
  booktitle = {arXiv preprint},
  year      = {2024}
}

@inproceedings{DAmbrosio2023TableTennis,
  author    = {D'Ambrosio, David B. and others},
  title     = {Robotic Table Tennis: A Case Study into a High Speed Learning System},
  booktitle = {Robotics: Science and Systems (RSS)},
  year      = {2023}
}

@inproceedings{Grandia2024Bipedal,
  author    = {Grandia, Ruben and Knoop, Espen and Hopkins, Michael A. and Wiedebach, Georg and Bishop, Jared and Pickles, Steven and M{\"u}ller, David and B{\"a}cher, Moritz},
  title     = {Design and Control of a Bipedal Robotic Character},
  booktitle = {Robotics: Science and Systems (RSS)},
  year      = {2024}
}

@misc{orsula2025spaceroboticsbench,
  title         = {Space Robotics Bench: Robot Learning Beyond Earth},
  author        = {Andrej Orsula and Matthieu Geist and Miguel Olivares-Mendez and Carol Martinez},
  year          = {2025},
  eprint        = {2509.23328},
  archivePrefix = {arXiv},
  primaryClass  = {cs.RO},
  url           = {https://arxiv.org/abs/2509.23328},
}

@misc{elhariry2025roboran,
      title={RoboRAN: A Unified Robotics Framework for Reinforcement Learning-Based Autonomous Navigation},
      author={Matteo El-Hariry and Antoine Richard and Ricard M. Castan and Luis F. W. Batista and Matthieu Geist and Cedric Pradalier and Miguel Olivares-Mendez},
      year={2025},
      eprint={2505.14526},
      archivePrefix={arXiv},
      primaryClass={cs.RO},
      url={https://arxiv.org/abs/2505.14526},
}

@article{wapman2021ssdt,
  author  = {Wapman, Jonathan D. and Sternberg, David C. and Lo, Kevin and Wang, Michael and Jones-Wilson, Laura L. and Mohan, Swati},
  title   = {Jet Propulsion Laboratory Small Satellite Dynamics Testbed Planar Air-Bearing Propulsion System Characterization},
  journal = {Journal of Spacecraft and Rockets},
  volume  = {58},
  number  = {4},
  pages   = {1126--1140},
  year    = {2021},
  doi     = {10.2514/1.A34857}
}

@inproceedings{bredenbeck2023reacsa,
  author    = {Bredenbeck, Anton and Vyas, Shubham and Suter, Willem and Zwick, Martin and Borrmann, Dorit and Olivares-Mendez, Miguel and N\"{u}chter, Andreas},
  title     = {{REACSA}: Actuated Floating Platform for Orbital Robotic Concept Testing and Control Software Development},
  booktitle = {Proc. ESA Symp. Advanced Space Technologies in Robotics and Automation (ASTRA)},
  year      = {2023}
}

@misc{stark2023linearmpc,
  author        = {Stark, Franek and Vyas, Shubham and Schildbach, Georg and Kirchner, Frank},
  title         = {Linear Model Predictive Control for a Planar Free-Floating Platform: A Comparison of Binary Input Constraint Formulations},
  year          = {2023},
  eprint        = {2312.10788},
  archivePrefix = {arXiv},
  primaryClass  = {eess.SY},
  url           = {https://arxiv.org/abs/2312.10788}
}

@article{floresabad2014review,
  author  = {Flores-Abad, Angel and Ma, Ou and Pham, Khanh and Ulrich, Steve},
  title   = {A Review of Space Robotics Technologies for On-Orbit Servicing},
  journal = {Progress in Aerospace Sciences},
  volume  = {68},
  pages   = {1--26},
  year    = {2014},
  doi     = {10.1016/j.paerosci.2014.03.002}
}

@article{moosavian2007freeflying,
  author  = {Moosavian, S. Ali A. and Papadopoulos, Evangelos},
  title   = {Free-flying Robots in Space: An Overview of Dynamics Modeling, Planning and Control},
  journal = {Robotica},
  volume  = {25},
  number  = {5},
  pages   = {537--547},
  year    = {2007},
  doi     = {10.1017/S0263574707003438}
}

@article{yoshida2003etsvii,
  author  = {Yoshida, Kazuya},
  title   = {Engineering Test Satellite {VII} Flight Experiments for Space Robot Dynamics and Control: Theories on Laboratory Test Beds Ten Years Ago, Now in Orbit},
  journal = {The International Journal of Robotics Research},
  volume  = {22},
  number  = {5},
  pages   = {321--335},
  year    = {2003},
  doi     = {10.1177/0278364903022005003}
}

@inproceedings{ogilvie2008orbitalexpress,
  author    = {Ogilvie, Andrew and Allport, Justin and Hannah, Michael and Lymer, John},
  title     = {Autonomous Satellite Servicing Using the Orbital Express Demonstration Manipulator System},
  booktitle = {Proc. 9th Int. Symp. Artificial Intelligence, Robotics and Automation in Space (i-SAIRAS)},
  year      = {2008}
}

@article{nagaoka2018dualarm,
  author  = {Nagaoka, Kenji and Kameoka, Ryota and Yoshida, Kazuya},
  title   = {Repeated Impact-Based Capture of a Spinning Object by a Dual-Arm Space Robot},
  journal = {Frontiers in Robotics and AI},
  volume  = {5},
  pages   = {115},
  year    = {2018},
  doi     = {10.3389/frobt.2018.00115}
}

@inproceedings{artigas2015oossim,
  author    = {Artigas, Jordi and De Stefano, Marco and Rackl, Wolfgang and Lampariello, Roberto and Brunner, Bernhard and Bertleff, Wieland and Burger, Robert and Porges, Oliver and Giordano, Alessandro Massimo and Borst, Christoph and Albu-Sch\"{a}ffer, Alin},
  title     = {The {OOS-SIM}: An On-Ground Simulation Facility for On-Orbit Servicing Robotic Operations},
  booktitle = {IEEE Int. Conf. Robotics and Automation (ICRA)},
  pages     = {2854--2860},
  year      = {2015},
  doi       = {10.1109/ICRA.2015.7139588}
}

@article{rybus2019planar,
  author  = {Rybus, Tomasz and Seweryn, Karol and Oles, Jakub and Basmadji, Fatina L. and Tarenko, Kamil and Moczydlowski, Rafal and Barcinski, Tomasz and Kindracki, Jan and Mezyk, Lukasz and Paszkiewicz, Pawel and Wolanski, Piotr},
  title   = {Application of a Planar Air-Bearing Microgravity Simulator for Demonstration of Operations Required for an Orbital Capture with a Manipulator},
  journal = {Acta Astronautica},
  volume  = {155},
  pages   = {211--229},
  year    = {2019},
  doi     = {10.1016/j.actaastro.2018.12.004}
}

@article{hovell2021deeprl,
  author  = {Hovell, Kirk and Ulrich, Steve},
  title   = {Deep Reinforcement Learning for Spacecraft Proximity Operations Guidance},
  journal = {Journal of Spacecraft and Rockets},
  volume  = {58},
  number  = {2},
  pages   = {254--264},
  year    = {2021},
  doi     = {10.2514/1.A34838}
}

@article{hovell2022capture,
  author  = {Hovell, Kirk and Ulrich, Steve},
  title   = {Laboratory Experimentation of Spacecraft Robotic Capture Using Deep-Reinforcement-Learning-Based Guidance},
  journal = {Journal of Guidance, Control, and Dynamics},
  volume  = {45},
  number  = {11},
  pages   = {2138--2146},
  year    = {2022},
  doi     = {10.2514/1.G006656}
}

@misc{chapin2025apiary,
  author        = {Chapin, Samantha and Stewart, Kenneth and Leontie, Roxana and Henshaw, Carl Glen},
  title         = {Autonomous Planning In-space Assembly Reinforcement-learning free-flYer ({APIARY}): International Space Station Astrobee Testing},
  year          = {2025},
  eprint        = {2512.03729},
  archivePrefix = {arXiv},
  primaryClass  = {cs.RO},
  url           = {https://arxiv.org/abs/2512.03729}
}

@misc{elhariry2025underactuated,
  author        = {El Hariry, Matteo and Cini, Andrea and Mellone, Giacomo and Balossino, Alessandro},
  title         = {Deep Reinforcement Learning Policies for Underactuated Satellite Attitude Control},
  year          = {2025},
  eprint        = {2505.00165},
  archivePrefix = {arXiv},
  primaryClass  = {eess.SY},
  url           = {https://arxiv.org/abs/2505.00165}
}

@article{gaudet2020terminal,
  author  = {Gaudet, Brian and Linares, Richard and Furfaro, Roberto},
  title   = {Terminal Adaptive Guidance via Reinforcement Meta-Learning: Applications to Autonomous Asteroid Close-Proximity Operations},
  journal = {Acta Astronautica},
  volume  = {171},
  pages   = {1--13},
  year    = {2020},
  doi     = {10.1016/j.actaastro.2020.02.036}
}

@article{retagne2024adaptive,
  author  = {Retagne, Wiebke and Dauer, Jonas and Waxenegger-Wilfing, G{\"u}nther},
  title   = {Adaptive Satellite Attitude Control for Varying Masses Using Deep Reinforcement Learning},
  journal = {Frontiers in Robotics and AI},
  volume  = {11},
  pages   = {1402846},
  year    = {2024},
  doi     = {10.3389/frobt.2024.1402846}
}

@article{hwangbo2019learning,
  author  = {Hwangbo, Jemin and Lee, Joonho and Dosovitskiy, Alexey and Bellicoso, Dario and Tsounis, Vassilios and Koltun, Vladlen and Hutter, Marco},
  title   = {Learning Agile and Dynamic Motor Skills for Legged Robots},
  journal = {Science Robotics},
  volume  = {4},
  number  = {26},
  pages   = {eaau5872},
  year    = {2019},
  doi     = {10.1126/scirobotics.aau5872}
}

@inproceedings{rudin2022learning,
  author    = {Rudin, Nikita and Hoeller, David and Reist, Philipp and Hutter, Marco},
  title     = {Learning to Walk in Minutes Using Massively Parallel Deep Reinforcement Learning},
  booktitle = {Proc. 5th Conf. on Robot Learning (CoRL)},
  series    = {Proceedings of Machine Learning Research},
  volume    = {164},
  pages     = {91--100},
  year      = {2022},
  url       = {https://proceedings.mlr.press/v164/rudin22a.html}
}

@inproceedings{makoviychuk2021isaacgym,
  author        = {Makoviychuk, Viktor and Wawrzyniak, Lukasz and Guo, Yunrong and Lu, Michelle and Storey, Kier and Macklin, Miles and Hoeller, David and Rudin, Nikita and Allshire, Arthur and Handa, Ankur and State, Gavriel},
  title         = {Isaac Gym: High Performance GPU-Based Physics Simulation for Robot Learning},
  booktitle     = {Proc. NeurIPS Track on Datasets and Benchmarks},
  year          = {2021},
  eprint        = {2108.10470},
  archivePrefix = {arXiv},
  url           = {https://arxiv.org/abs/2108.10470}
}

@article{mittal2023orbit,
  author  = {Mittal, Mayank and Yu, Calvin and Yu, Qinxi and Liu, Jingzhou and Rudin, Nikita and Hoeller, David and Yuan, Jia Lin and Singh, Ritvik and Guo, Yunrong and Mazhar, Hammad and Mandlekar, Ajay and Babich, Buck and State, Gavriel and Hutter, Marco and Garg, Animesh},
  title   = {Orbit: A Unified Simulation Framework for Interactive Robot Learning Environments},
  journal = {IEEE Robotics and Automation Letters},
  volume  = {8},
  number  = {6},
  pages   = {3740--3747},
  year    = {2023},
  doi     = {10.1109/LRA.2023.3270034}
}

@article{kaufmann2023champion,
  author  = {Kaufmann, Elia and Bauersfeld, Leonard and Loquercio, Antonio and M{\"u}ller, Matthias and Koltun, Vladlen and Scaramuzza, Davide},
  title   = {Champion-Level Drone Racing Using Deep Reinforcement Learning},
  journal = {Nature},
  volume  = {620},
  number  = {7976},
  pages   = {982--987},
  year    = {2023},
  doi     = {10.1038/s41586-023-06419-4}
}

@article{mittal2025isaaclab,
  title={Isaac Lab: A GPU-Accelerated Simulation Framework for Multi-Modal Robot Learning},
  author={Mayank Mittal et al.},
  journal={arXiv preprint arXiv:2511.04831},
  year={2025},
  url={https://arxiv.org/abs/2511.04831}
}

@article{schwarke2025rslrl,
  title={RSL-RL: A Learning Library for Robotics Research},
  author={Schwarke, Clemens and Mittal, Mayank and Rudin, Nikita and Hoeller, David and Hutter, Marco},
  journal={arXiv preprint arXiv:2509.10771},
  year={2025}
}

@misc{newway150mmairbearing,
  author       = {{New Way Air Bearings}},
  title        = {150mm Flat Round Air Bearing (SKU: S1015001)},
  howpublished = {\url{https://www.newwayairbearings.com/catalog/product/150mm-flat-round-air-bearings/}},
  year         = {2026},
}

@misc{odriverobotics,
  author       = {{ODrive Robotics}},
  title        = {ODrive: High-Performance Motor Control},
  howpublished = {\url{https://odriverobotics.com/}},
  year         = {2026},
  note         = {Accessed: July 20, 2026}
}

@misc{schulman2017proximalpolicyoptimizationalgorithms,
      title={Proximal Policy Optimization Algorithms}, 
      author={John Schulman and Filip Wolski and Prafulla Dhariwal and Alec Radford and Oleg Klimov},
      year={2017},
      eprint={1707.06347},
      archivePrefix={arXiv},
      primaryClass={cs.LG},
      url={https://arxiv.org/abs/1707.06347}, 
}

@article{kalman1960contributions,
  author  = {K{\'a}lm{\'a}n, Rudolf E.},
  title   = {Contributions to the Theory of Optimal Control},
  journal = {Bolet{\'i}n de la Sociedad Matem{\'a}tica Mexicana},
  volume  = {5},
  number  = {1},
  pages   = {102--119},
  year    = {1960}
}

\appendix
\subsection{Reproducibility and Open Access}
To support reproducibility and lower the barrier to entry for learning-based spacecraft GNC research, we openly release all assets associated with this work. The complete hardware design (CAD and BOM), firmware, ROS 2 packages, simulation environments, training scripts, experimental logs, and supplementary videos are available at \url{https://snt-spacer.github.io/UniLuFP/}.

\subsection{More Implementation Details}
\label{sec:appendix:dr}

All policies are trained with Proximal Policy Optimization (PPO), using the RSL-RL \cite{schwarke2025rslrl} implementation within IsaacLab~\cite{mittal2025isaaclab}, across 5 random seeds, with 4096 parallel environments during training and 640 environments at evaluation. We consider two policy architectures: a feed-forward multi-layer perceptron (MLP) and a recurrent variant that inserts a single-layer gated recurrent unit (GRU) before the actor and critic heads to provide memory for rejecting the unobserved, time-varying disturbances induced by arm motion. The two architectures share all optimization hyperparameters; they differ only in the recurrent core and in the training budget (the recurrent policy is trained for twice as many iterations to accommodate the additional parameters). \Cref{tab:hyperparameters} summarizes both configurations.

\begin{table}[!h]
    \centering
    \vspace{-0.05in}
    \caption{PPO hyperparameters for the Dynamic Disturbance Rejection policies, for the feed-forward (MLP) and recurrent (GRU) variants. Shared values span both columns.}
    \vspace{-0.05in}
    \begin{tabular}{lcc}
    \toprule
    \textbf{Hyperparameter} & \textbf{PPO (MLP)} & \textbf{PPO (GRU)} \\

    \midrule
    Number of envs & \multicolumn{2}{c}{4096} \\
    Number of steps per iteration & \multicolumn{2}{c}{16} \\
    Number of learning epochs & \multicolumn{2}{c}{5} \\
    Number of mini-batches & \multicolumn{2}{c}{4} \\
    Clip range & \multicolumn{2}{c}{0.2} \\
    Entropy coefficient & \multicolumn{2}{c}{0.005} \\
    Value loss coefficient & \multicolumn{2}{c}{1.0} \\
    GAE balancing factor $\lambda$ & \multicolumn{2}{c}{0.95} \\
    Discount factor $\gamma$ & \multicolumn{2}{c}{0.99} \\
    Desired KL-divergence & \multicolumn{2}{c}{0.01} \\
    Initial learning rate & \multicolumn{2}{c}{1$\times$10$^{-3}$} \\
    Learning-rate schedule & \multicolumn{2}{c}{Adaptive} \\
    Max gradient norm & \multicolumn{2}{c}{1.0} \\
    Initial action noise std & \multicolumn{2}{c}{1.0} \\
    Actor and critic MLP & \multicolumn{2}{c}{[64, 64]} \\
    Activation function & \multicolumn{2}{c}{Tanh} \\
    \midrule
    Recurrent core & --- & GRU \\
    GRU hidden dimension & --- & 64 \\
    GRU layers & --- & 1 \\
    Training iterations & 1000 & 2000 \\
    \bottomrule
    \end{tabular}
    \label{tab:hyperparameters}
\end{table}

\paragraph{Domain randomization}
To bridge the sim-to-real gap, a set of per-environment physical parameters of the platform base is re-sampled at every episode reset. The \emph{base mass} is perturbed by a uniform offset in $[-5, +5]$\,kg. The \emph{center of mass} is displaced by up to $0.1$\,m in a uniformly random horizontal direction. A \emph{constant external wrench} is drawn once per episode and held for its entire duration: a planar bias force of up to $1$\,N along a random direction, together with a bias yaw torque of up to $0.05$\,N\,m, emulating unmodeled effects such as air-bearing tilt, residual table slope, and thruster imbalance. These three are the randomizations enabled during training; action-level noise and thruster-gain rescaling are implemented in the framework but disabled for the reported experiments. The platform's moment of inertia is not randomized explicitly; inertial variation is instead induced physically through the three static \levion arm configurations (\cref{fig:arms-mode}) and implicitly through the mass and CoM randomization.

\Cref{table:reward_functions} details the reward terms used to train each policy. Every reward is the weighted sum of a task objective and a set of regularization terms that shape \emph{how} the objective is achieved. For \textbf{Point-to-Pose Navigation} (a), the dominant term is an exponential pose reward that rewards simultaneous convergence in position ($d_p$) and heading ($d_h$); this is complemented by soft linear- and angular-velocity penalties that discourage overshoot, a boundary term that keeps the platform inside the workspace, and the platform regularization of block (c). The latter is a set of small penalties on the actuation itself---joint acceleration, thruster action rate and effort, reaction-wheel speed and torque, and arm motion and collisions---that together suppress chattering, discourage wasteful actuation, and yield smooth commands. As noted in the table, the reaction-wheel and arm penalties are active only for the actuator subsets that include those effectors.

The \textbf{Reaction-Wheel Stabilization} task (b) is the momentum-damping experiment: the platform spawns with a residual body spin of $0.1$--$0.5$\,rad/s and must null its yaw rate using only the reaction wheel. The single task term rewards driving the yaw rate $\omega_z$ toward zero ($\omega_z^{\star}=0$) through a sharp exponential kernel ($\lambda_{\omega}=0.5$), so the reward grows steeply as the platform comes to rest. Because this policy actuates only the reaction wheel, its regularization is wheel-specific rather than the terms in (c): a quadratic penalty on internal wheel speed $\omega_{\text{rw}}$ discourages saturation (a saturated wheel produces no usable torque), and a penalty on commanded torque $\tau_{\text{rw}}$ discourages unnecessary actuation, encouraging the policy to brake decisively and then desaturate. These two penalties are deliberately small relative to the unit-weight stabilization term, so they regularize the behavior without competing with the primary objective.

\subsection{More Experiments}

\begin{table}[h]
    \centering
    \setlength{\tabcolsep}{7pt}
    \renewcommand{\arraystretch}{1.2}
    \resizebox{0.48\textwidth}{!}{
        \begin{tabular}{l cc}
        \toprule
        \textbf{Method} & $e_p\,[\text{m}]\downarrow$ & $e_o\,[\text{rad}]\downarrow$ \\
        \midrule
        \ourrow \multicolumn{3}{l}{\textbf{(a) Point-to-Pose Navigation --- all actuators}} \\
        \cdashline{1-3}\noalign{\vskip 0.6mm}
        PPO (Sim)                            & 0.0022 \ci{0.0015} & 0.0044 \ci{0.0033} \\
        PPO + DR (Sim)                       & 0.0274 \ci{0.0760} & 0.0192 \ci{0.0187} \\
        PPO + DR (Real)                      & 0.0584 \ci{0.0412} & 0.0560 \ci{0.0151} \\
        PPO-GRU + DR, 1000\,ep (Sim)         & 0.0209 \ci{0.0344} & 0.0277 \ci{0.0255} \\
        PPO-GRU + DR, 1000\,ep (Real, chkpt. 1) & 0.0862 \ci{0.0862} & 0.0443 \ci{0.0307} \\
        PPO-GRU + DR, 1000\,ep (Real, chkpt. 2) & 0.0855 \ci{0.0161} & 0.0211 \ci{0.0173} \\
        PPO-GRU + DR, 2000\,ep (Sim)         & 0.0124 \ci{0.0191} & 0.0129 \ci{0.0124} \\
        PPO-GRU + DR, 2000\,ep (Real, chkpt. 1) & 0.0214 \ci{0.0484} & 0.0484 \ci{0.0239} \\
        PPO-GRU + DR, 2000\,ep (Real, chkpt. 2) & 0.0081 \ci{0.0006} & 0.0257 \ci{0.0146} \\
        \midrule
        \ourrow \multicolumn{3}{l}{\textbf{(b) Dynamic Disturbance Rejection}} \\
        \cdashline{1-3}\noalign{\vskip 0.6mm}
        PPO (Sim)                            & 0.012 \ci{0.002} & 0.016 \ci{0.003} \\
        PPO + DR (Sim)                       & 0.029 \ci{0.003} & 0.033 \ci{0.004} \\
        PPO + DR (Real, seed 1)              & 0.1346 \ci{0.0405} & 0.0438 \ci{0.0361} \\
        PPO + DR (Real, seed 3)              & 0.1502 \ci{0.0292} & 0.0708 \ci{0.0184} \\
        PPO-GRU (Sim)                        & 0.0070 \ci{0.0054} & 0.0074 \ci{0.0108} \\
        PPO-GRU + DR (Sim)                   & 0.0156 \ci{0.0161} & 0.0340 \ci{0.0574} \\
        PPO-GRU + DR (Real, seed 4)          & 0.1521 \ci{0.0441} & 0.0887 \ci{0.0741} \\
        PPO-GRU + DR (Real, seed 1)          & 0.0889 \ci{0.0405} & 0.0346 \ci{0.0179} \\
        \bottomrule
        \end{tabular}
    }
    \caption{\textbf{All-actuator navigation and dynamic disturbance rejection --- full ablation.} Final position error $e_p$ and heading error $e_o$ for the two studies in which the arms are actuated (all-actuators navigation) or continuously animated as a disturbance (dynamic rejection), so no static arm pose applies. For the all-actuators policy we also sweep the recurrent training budget (1000 vs.\ 2000 epochs); for the hardware entries we report multiple training seeds / deployment runs. Lower is better.}
    \label{tab:results_ablation_allact_dyn}
    \vspace{-1em}
\end{table}

\Cref{tab:results_pose_nav_full,tab:results_ablation_allact_dyn} report the complete ablation study underlying the curated hardware numbers of \cref{sec:experiments}. Across tasks we sweep up to four axes. \textbf{Arm configuration} varies the platform's effective moment of inertia by fixing the \levion arms at one of the three static poses of \cref{fig:arms-mode}: \emph{Side}, \emph{Rest}, and \emph{Closed}. This axis applies only to the reduced-actuator subsets (thrusters only, thrusters~+~reaction wheel); the all-actuators navigation policy and the disturbance-rejection policy either control or continuously animate the arms and therefore have no single static pose. \textbf{Policy architecture} contrasts the feed-forward PPO (MLP) policy with the recurrent PPO-GRU policy, whose hidden state provides the memory needed to infer unobserved inertial shifts. \textbf{Domain randomization} (DR) toggles randomization of the base mass, center of mass, and a constant external wrench (a per-episode bias force and torque) during training, and is a prerequisite for hardware transfer; the exact ranges are given in \cref{sec:appendix:dr}. \textbf{Deployment} distinguishes simulation (Sim) from hardware (Real); for the hardware entries we additionally report several independent training seeds / deployment runs to expose the run-to-run spread of sim-to-real transfer.

\begin{table*}[h]
    \centering
    \caption{Reward terms used to train the learned control policies. Each block lists a task-specific reward, given as the weighted sum of its rows. Block (c) collects the platform-regularization penalties added to the \emph{Point-to-Pose Navigation} policy (a); as noted per row, individual terms apply only to configurations that include the corresponding actuators (reaction wheel, arms). The reaction-wheel \emph{Stabilization} policy (b) actuates only the wheel and is therefore regularized by just the reaction-wheel saturation and usage terms, which share the definitions and weights of the matching rows in (c). Distances $d_\bullet$ denote the absolute error between the current and target quantity; negative weights indicate penalties.}
    \renewcommand{\arraystretch}{1.05} 
    \resizebox{1\linewidth}{!}{
    \begin{tabular}{l l l l l}
    \toprule 
    Term & Expression & Weight & Scalar & Description \\
        
    \midrule 
    \noalign{\vskip -0.2mm}
      \ourrow \textbf{(a) Point-to-Pose Navigation}  & $r_t^{\text{point-pose-nav}} = r_t^{\text{pose}} + r_t^{v,\omega} + r_t^{\text{bnd}}$ & & & \\
      
      \noalign{\vskip 0.4mm}\cdashline{1-5}\noalign{\vskip 0.8mm}
        Pose error & $\alpha_{\text{pose}} (\text{exp}(-d_p / \lambda_p) \times \text{exp}(-d_h / \lambda_h))$ & $\alpha_{\text{pose}}=1.0$ & $\lambda_p,\lambda_h = 1.0$ & Precise spatial convergence and alignment to target.\\

        Linear Velocity & $\alpha_{\text{v}} \text{clip}(v-v_{\text{min}}, 0, v_{\text{max}}-v_{\text{min}})$ & $\alpha_{\text{v}}=-0.05$ & $v_{\text{min}}=0.5, v_{\text{max}}=2.0$ & Velocity regulation and movement encouragement.\\

        Angular Velocity & $\alpha_{\omega} \text{clip}(\omega-\omega_{\text{min}}, 0, \omega_{\text{max}}-\omega_{\text{min}})$ & $\alpha_{\omega}=-0.05$ & $\omega_{\text{min}}=0.5, \omega_{\text{max}}=20.0$ & Control of rotational stability.\\

        Boundary & $\alpha_{\text{bnd}} \text{exp}(-d_b / \lambda_b)$ & $\alpha_{\text{bnd}}=-10.0$ & $\lambda_b=1.0$ & Safety constraint to stay within workspace.\\

    \midrule 
    \noalign{\vskip -0.2mm}
      \ourrow \textbf{(b) Reaction-Wheel Stabilization (Momentum Damping)}  & $r_t^{\text{stab}} = r_t^{\omega} + r_t^{\text{sat}} + r_t^{\text{use}}$ & & & \\

      \noalign{\vskip 0.4mm}\cdashline{1-5}\noalign{\vskip 0.8mm}
        Angular-rate nulling & $\alpha_{\omega} \text{exp}(-|\omega_z - \omega_z^{\star}| / \lambda_{\omega})$ & $\alpha_{\omega}=1.0$ & $\lambda_{\omega}=0.5,\ \omega_z^{\star}=0$ & Rewards driving the platform yaw rate $\omega_z$ to zero, i.e.\ dissipating the initial angular momentum.\\

        Reaction-wheel saturation & $\gamma_{\text{sat}} \, \omega_{\text{rw}}^2$ & $\gamma_{\text{sat}}=-2.5\times 10^{-6}$ & & Penalizes high internal wheel speed $\omega_{\text{rw}}$ to keep the wheel from saturating and preserve control authority.\\

        Reaction-wheel usage & $\gamma_{\text{use}} \, |\tau_{\text{rw}}|$ & $\gamma_{\text{use}}=-0.05$ & & Penalizes the commanded wheel torque $\tau_{\text{rw}}$ to discourage unnecessary actuation.\\

    \midrule 
    \noalign{\vskip -0.2mm}
      \ourrow \textbf{(c) Platform Regularization (Navigation)}  & $r_t^{\text{robot}} = r^{\text{acc}} + r^{\text{rate}} + r^{\text{eff}} + r^{\text{sat}} + r^{\text{use}} + r^{\text{arm}} + r^{\text{col}}$ & & & \\

      \noalign{\vskip 0.4mm}\cdashline{1-5}\noalign{\vskip 0.8mm}
        Joint acceleration & $\gamma_{\text{acc}} \sum_j \ddot{q}_j^2$ & $\gamma_{\text{acc}}=-2.5\times 10^{-6}$ & & Penalizes summed squared joint accelerations for smoother motion. \emph{(all configs)}\\

        Thruster action rate & $\gamma_{\text{rate}} \sum_i |a_t^i - a_{t-1}^i|$ & $\gamma_{\text{rate}}=-0.015$ & & Penalizes step-to-step change in thruster commands to reduce chattering. \emph{(all configs)}\\

        Thruster effort & $\gamma_{\text{eff}} \sum_i |f_i| / f_{\max}$ & $\gamma_{\text{eff}}=-0.01$ & $f_{\max}=1.0$ & Penalizes total normalized thrust to reduce sustained actuation. \emph{(all configs)}\\

        Reaction-wheel saturation & $\gamma_{\text{sat}} \, \omega_{\text{rw}}^2$ & $\gamma_{\text{sat}}=-2.5\times 10^{-6}$ & & Penalizes high wheel speed to avoid saturation. \emph{(configs with reaction wheel)}\\

        Reaction-wheel usage & $\gamma_{\text{use}} \, |\tau_{\text{rw}}|$ & $\gamma_{\text{use}}=-0.05$ & & Penalizes commanded wheel torque. \emph{(configs with reaction wheel)}\\

        Arm action rate & $\gamma_{\text{arm}} \sum_j |\theta_t^j - \theta_{t-1}^j|$ & $\gamma_{\text{arm}}=-0.1$ & & Penalizes step-to-step change in arm position targets to limit induced vibration. \emph{(all-actuators config)}\\

        Arm collision & $\gamma_{\text{col}} \, \mathbbm{1}[F_{\text{arm}} > F_{\text{thr}}]$ & $\gamma_{\text{col}}=-1.0$ & $F_{\text{thr}}=0.1$\,N & Penalizes contact detected on the arm links. \emph{(all-actuators config)}\\

    \bottomrule
    \end{tabular}}
    \vspace{-0.1in}
    \label{table:reward_functions} 
\end{table*}

\begin{table*}[t]
    \centering
    \setlength{\tabcolsep}{6pt}
    \renewcommand{\arraystretch}{1.2}
    \caption{\textbf{Point-to-Pose Navigation - static-arm ablation.} Final position error $e_p$ and heading error $e_o$ for the reduced-actuator subsets, evaluated across the three static \levion arm poses (\emph{Side}, \emph{Rest}, \emph{Closed}) of \cref{fig:arms-mode}. Each subset is ablated over policy architecture (feed-forward PPO vs.\ recurrent PPO-GRU), domain randomization (DR), and deployment (Sim vs.\ Real hardware). Entries are mean~$\pm$~std over evaluation episodes; lower is better.}
    \begin{tabular}{l cc cc cc}
        \toprule
        & \multicolumn{2}{c}{\textit{Side}} & \multicolumn{2}{c}{\textit{Rest}} & \multicolumn{2}{c}{\textit{Closed}} \\
        \cmidrule(lr){2-3}\cmidrule(lr){4-5}\cmidrule(lr){6-7}
        \textbf{Method} & $e_p\,[\text{m}]\downarrow$ & $e_o\,[\text{rad}]\downarrow$ & $e_p\,[\text{m}]\downarrow$ & $e_o\,[\text{rad}]\downarrow$ & $e_p\,[\text{m}]\downarrow$ & $e_o\,[\text{rad}]\downarrow$ \\
        \midrule
        \ourrow \multicolumn{7}{l}{\textbf{(a) Thrusters and reaction wheel}} \\
        \cdashline{1-7}\noalign{\vskip 0.6mm}
        PPO (Sim)           & 0.0021 \ci{0.0010} & 0.0038 \ci{0.0043} & 0.0031 \ci{0.0026} & 0.0020 \ci{0.0022} & 0.0034 \ci{0.0023} & 0.0018 \ci{0.0013} \\
        PPO + DR (Sim)      & 0.0155 \ci{0.0423} & 0.0193 \ci{0.0175} & 0.0204 \ci{0.0624} & 0.0406 \ci{0.0513} & 0.0154 \ci{0.0360} & 0.0122 \ci{0.0122} \\
        PPO + DR (Real)     & 0.0115 \ci{0.0076} & 0.0302 \ci{0.0088} & 0.0246 \ci{0.0155} & 0.0313 \ci{0.0139} & 0.0406 \ci{0.0227} & 0.0121 \ci{0.0066} \\
        PPO-GRU (Sim)       & 0.0137 \ci{0.0112} & 0.0047 \ci{0.0039} & 0.0220 \ci{0.0217} & 0.0047 \ci{0.0036} & 0.0045 \ci{0.0055} & 0.0040 \ci{0.0033} \\
        PPO-GRU + DR (Sim)  & 0.0083 \ci{0.0090} & 0.0081 \ci{0.0085} & 0.0097 \ci{0.0098} & 0.0069 \ci{0.0059} & 0.0145 \ci{0.0111} & 0.0067 \ci{0.0072} \\
        PPO-GRU + DR (Real) & 0.0547 \ci{0.0422} & 0.0275 \ci{0.0154} & 0.0482 \ci{0.0271} & 0.0279 \ci{0.0204} & 0.0679 \ci{0.0500} & 0.0598 \ci{0.0345} \\
        \midrule
        \ourrow \multicolumn{7}{l}{\textbf{(b) Thrusters only}} \\
        \cdashline{1-7}\noalign{\vskip 0.6mm}
        PPO (Sim)           & 0.0025 \ci{0.0024} & 0.0028 \ci{0.0024} & 0.0047 \ci{0.0052} & 0.0028 \ci{0.0035} & 0.0047 \ci{0.0045} & 0.0027 \ci{0.0022} \\
        PPO + DR (Sim)      & 0.0080 \ci{0.0135} & 0.0086 \ci{0.0071} & 0.0189 \ci{0.0727} & 0.0104 \ci{0.0173} & 0.0124 \ci{0.0452} & 0.0087 \ci{0.0088} \\
        PPO + DR (Real)     & 0.0087 \ci{0.0067} & 0.0202 \ci{0.0129} & 0.0176 \ci{0.0198} & 0.0160 \ci{0.0063} & 0.0348 \ci{0.0126} & 0.0310 \ci{0.0413} \\
        PPO-GRU (Sim)       & 0.0064 \ci{0.0106} & 0.0024 \ci{0.0026} & 0.0057 \ci{0.0089} & 0.0046 \ci{0.0032} & 0.0099 \ci{0.0111} & 0.0055 \ci{0.0041} \\
        PPO-GRU + DR (Sim)  & 0.0069 \ci{0.0087} & 0.0087 \ci{0.0082} & 0.0056 \ci{0.0065} & 0.0105 \ci{0.0081} & 0.0075 \ci{0.0088} & 0.0062 \ci{0.0051} \\
        PPO-GRU + DR (Real) & 0.0241 \ci{0.0160} & 0.0335 \ci{0.0155} & 0.0745 \ci{0.0396} & 0.0219 \ci{0.0028} & 0.0527 \ci{0.0192} & 0.0414 \ci{0.0363} \\
        \bottomrule
    \end{tabular}
        \label{tab:results_pose_nav_full}
\end{table*}

Two trends emerge from \cref{tab:results_pose_nav_full}. First, in simulation both architectures reach sub-centimeter position error in every arm configuration; adding DR raises the simulation error only marginally while enabling hardware transfer, the expected robustness--accuracy trade-off. Second, the arm configuration has a comparatively small effect on the learned controllers relative to the sim-to-real gap, confirming that the policies absorb the inertia change induced by the static arm pose rather than overfitting to a particular one. On hardware, the feed-forward PPO+DR policy transfers on par with or slightly better than PPO-GRU+DR for this \emph{static} navigation task, consistent with the recurrent policy being most valuable when the disturbance is time-varying (\cref{tab:results_ablation_allact_dyn}) rather than fixed.

For the all-actuators and disturbance-rejection studies (\cref{tab:results_ablation_allact_dyn}), the recurrent training-length sweep shows that the longer 2000-epoch schedule yields the best-transferring all-actuators policy ($e_p=0.0081$\,m on hardware). Under the continuous CoM disturbance the recurrent policy clearly outperforms the feed-forward one in simulation ($e_p=0.0070$ vs.\ $0.012$\,m), and its best hardware seed attains the lowest real-world position error ($e_p=0.0889$\,m), motivating the recurrent policy for the dynamic task. The larger hardware position errors in this task, relative to static navigation, reflect the never-settling nature of the injected disturbance discussed in \cref{sec:experiments}.

\end{document}